\documentclass[lettersize,journal]{IEEEtran}
\usepackage{amsmath,amsfonts}
\usepackage{algorithmic}
\usepackage{algorithm}
\usepackage{array}
\usepackage[caption=false,font=normalsize,labelfont=sf,textfont=sf]{subfig}
\usepackage{textcomp}
\usepackage{stfloats}
\usepackage{url}
\usepackage{verbatim}
\usepackage{graphicx}
\usepackage{cite}
\usepackage{multirow}
\usepackage{tabularx}
\usepackage{booktabs}
\usepackage{xcolor}
\usepackage{colortbl}
\usepackage{amssymb}
\usepackage{diagbox}
\usepackage{makecell}
\definecolor{deepgreen}{rgb}{0.0, 0.5, 0.0}
\definecolor{deepred}{rgb}{0.5, 0.0, 0.0}

\newcommand{\cmark}{\textcolor{deepgreen}{\checkmark}}
\newcommand{\xmark}{\textcolor{deepred}{\texttimes}}

\renewcommand{\arraystretch}{1.5}
\newcolumntype{C}[1]{>{\centering\arraybackslash}m{#1}}

\begin{document}

\title{LuSNAR:A Lunar Segmentation, Navigation and Reconstruction Dataset based on Multi-sensor for Autonomous Exploration}

\author{Jiayi Liu,~Qianyu Zhang,~Xue Wan,~Shengyang Zhang,~Yaolin Tian,\\Haodong Han,~Yutao Zhao,~Baichuan Liu,~Zeyuan Zhao,~Xubo Luo
\thanks{This work was supported  by the National Natural Science Foundation of China (No.42171445).\textit{(Corresponding author: Xue Wan.)}}
\thanks{
    All authors are with the University of Chinese Academy of Sciences, Beijing 101408, China, 
    and also with the Technology and Engineering Center for Space Utilization, 
    Chinese Academy of Sciences, Beijing 100094, China. 
    (e-mail: \{liujiayi21, zhangqianyu22, zhangshengyang22, tianyaolin21, hanhaodong23, 
    zhaoyutao22, liubaichuan23, zhaozeyuan23, luoxubo23\}@mails.ucas.ac.cn; 
    wanxue@csu.ac.cn).
}
}

\markboth{IEEE TRANSACTIONS ON GEOSCIENCE AND REMOTE SENSING}%
{Shell \MakeLowercase{\textit{et al.}}: A Sample Article Using IEEEtran.cls for IEEE Journals}


\maketitle

\begin{abstract}
With the complexity of lunar exploration missions, the moon needs to have a higher level of autonomy. Environmental perception and navigation algorithms are the foundation for lunar rovers to achieve autonomous exploration. The development and verification of algorithms require highly reliable data support. Most of the existing lunar datasets are targeted at a single task, lacking diverse scenes and high-precision ground truth labels. To address this issue, we propose a multi-task, multi-scene, and multi-label lunar benchmark dataset LuSNAR. This dataset can be used for comprehensive evaluation of autonomous perception and navigation systems, including high-resolution stereo image pairs, panoramic semantic labels, dense depth maps, LiDAR point clouds, and the position of rover. In order to provide richer scene data, we built 9 lunar simulation scenes based on Unreal Engine. Each scene is divided according to topographic relief and the density of objects. To verify the usability of the dataset, we evaluated and analyzed the algorithms of semantic segmentation, 3D reconstruction, and autonomous navigation. The experiment results prove that the dataset proposed in this paper can be used for ground verification of tasks such as autonomous environment perception and navigation, and provides a lunar benchmark dataset for testing the accessibility of algorithm metrics. We make LuSNAR publicly available at: https://github.com/zqyu9/LuSNAR-dataset. 
\end{abstract}

\begin{IEEEkeywords}
Autonomous exploration, Semantic segmentation, Navigation and localization, 3D Reconstruction.
\end{IEEEkeywords}

\section{Introduction}
\IEEEPARstart{T}{he}rovers, as planetary robots, play a crucial role in facilitating human exploration of extraterrestrial celestial bodies and enhancing our understanding of the universe. As the nearest celestial body to the earth, lunar exploration tasks attracts scientists for its geological evolution and internal structure. After the peak of lunar exploration led by the United States and former Soviet Union in 1976, space-faring nations such as China and India have successively sent satellites and rovers to the Moon for scientific exploration~\cite{gao2017review}. China successfully launched the Yutu and Yutu-2 rovers in Chang’E-3 (CE-3) and s Chang’e-4 (CE-4) mission, enabling exploration and investigation on the moon~\cite{xiao2015young,li2021overview}.In August of 2023, India's Chandrayaan-3 achieved a successful soft landing on the moon, thus becoming the fourth nation for lunar exploration. 

With the increasing complexity of deep space exploration missions, the long communication chain between Earth and the moon greatly restricts the human-in-loop operation and control of rovers~\cite{frank2016developing}.The safety and real-time performance of future exploration tasks require increasing autonomous ability of the rovers. Autonomous environmental perception and navigation are the foundation to ensure the safety and efficiency of lunar exploration tasks. The development and validation of perception and navigation algorithms demand a substantial amount of data with various scenes includes different topography and object distributions. In particular, deep learning algorithms rely on datasets with ground truth labels for model training and testing~\cite{minar2018recent}. Compared to autonomous driving tasks on Earth, the lunar surface environment presents a scenario characterized by unstructured terrain, minimal texture, and sparse features, making it challenging for lunar rovers to acquire diverse environmental data. This presents difficulties for the rovers in terms of terrain perception, localization, and mapping within the lunar environment. High-quality ground truth labels and diverse scene data can enhance the accuracy and generalizability of the model. Thus, a lunar benchmark dataset supporting the tasks of environmental perception and navigation will serve as a testbed for the comparison and evaluation of various algorithms.

Driven by the new wave of lunar exploration, some datasets have been proposed focused on the navigation and recognition on lunar surface. Furgale et al.~\cite{furgale2012devon} proposed Devon Island dataset for lunar terrain navigation. The sensors include stereo images, 3D laser ranging scans, and location data collected on terrain on Earth where topographic features are similar to the moon. Vayugundla et al.~\cite{vayugundla2018datasets} proposed LRNT dataset based on data collected by the Lightweight Rover Unit (LRU), and can be used to evaluate rover navigation. The S3LI dataset~\cite{giubilato2022challenges} and the LRNT dataset were collected at the same location and are also employed to validate and evaluate visual-inertial SLAM. Roman et al.~\cite{Romain2019} proposed a lunar landscape simulation dataset containing semantic labels for sky, smaller rocks and larger rocks, which can be used for training and testing semantic segmentation algorithms for lunar scene. These datasets show potential application value for future lunar semantic perception, positioning and navigation missions. However, the lack of scene diversity makes it difficult to evaluate the real performance of the algorithms, and moreover, the generalization of trained model can hardly be fulfilled. Therefore, the existing dataset is difficult to support the diverse needs of future lunar exploration. 

The autonomous exploration of rovers involves the collaborative execution of multiple tasks, including semantic perception, obstacle recognition, path planning, navigation and positioning, and terrain reconstruction. These tasks are connected and influenced each other. The dataset merely focused on single task cannot provide a comprehensive performance evaluation of the autonomous exploration system. For example, the failure detection of one stone in the semantic segmentation will lead to errors in obstacle map generation, and thus the path planning may be inaccurate; Incorrect navigation poses used for multi-site point cloud stitching can significantly impact the accuracy and reliability of path planning, potentially leading to unsafe driving for rovers. Datasets for a single task can only be used to validate and optimize algorithms for specific tasks, but cannot be used to comprehensively evaluate perception and navigation systems. The dataset must be combined with multi-task modules to achieve efficient and robust autonomous exploration.

Autonomous exploration missions require lunar rovers to adapt to diverse and unknown lunar environments, which may include vast plains, rocky terrain with dense distribution, and rugged crater bottoms which may have different distribution training data on Earth. The generalization of the environment perception algorithms is important. However, this is hard to achieve because the data collection scenario for real lunar exploration missions is limited, and relying solely on real exploration data for training and testing cannot improve the generalization of algorithms. Another potential solution is to establish lunar surface similar scenarios on Earth. However, it is difficult to build large number of lunar simulation scenarios with different topography features. A single-scene dataset may make the algorithm only applicable to lunar surface scenes under specific conditions, resulting in poor performance in large range explorations. Especially for deep learning algorithms that rely on a large amount of training data, using only samples collected in the same scenario may lead to overfitting of the model. Therefore, in order to improve the generalization ability of the algorithm, the dataset is necessary to cover different terrains and landforms from the data level.

Another challenge is the collection of a large amount of ground truth labels with high-quality and less labor-intensive. The existing dataset uses estimated values or manual annotations as labels, lacking reliable high-precision ground truth (depth information, semantic maps, pose). The cost of manual annotation is high, and its reliability is poor. Low-quality ground truth labels contain noise and error information, which can be mistaken by the model as true features, leading to the learning of incorrect rules. High-precision ground truth labels assist algorithms in learning and predicting more effectively, and accurately evaluate the performance of models.

\begin{figure*}[!t]
\centering
\includegraphics[width=\textwidth]{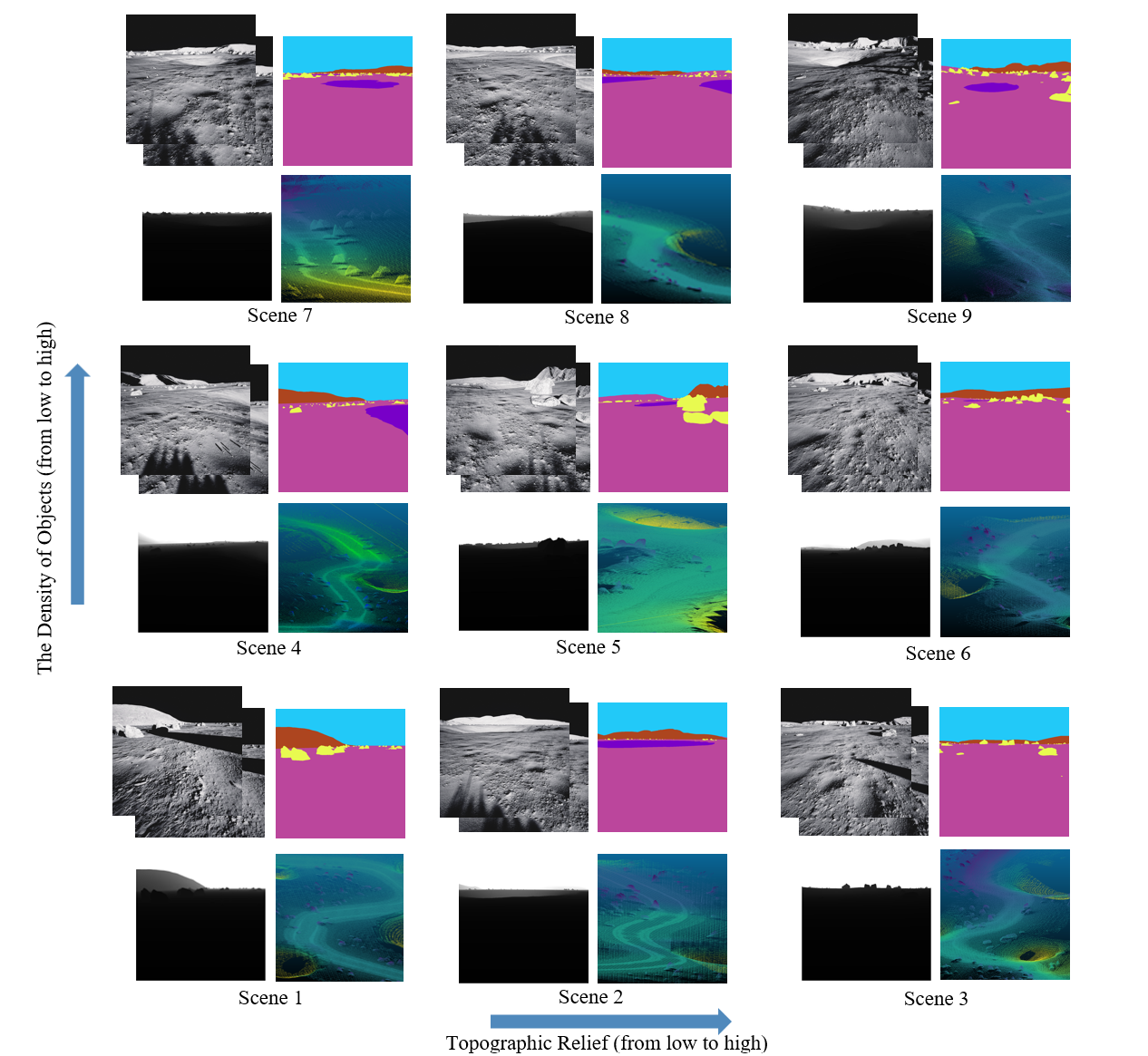}
\caption{Overview of 9 lunar surface scenes with different topographic relief and density of objects.}
\label{figure_1}
\end{figure*}

To address the above needs and challenges, this paper proposed a benchmark dataset named LuSNAR including nine simulated lunar surface scenes based on Unreal Engine 4, and it is different from many existing benchmark datasets which only focus on a single task, such as navigation or recognition. LuSNAR is the first lunar scene benchmark dataset that provides high-resolution stereo image pairs, panoramic semantic labels, depth maps, point clouds, poses, and timestamps. This dataset will encourage researchers to investigate potential perception, navigation and reconstruction for lunar exploration based on multiple sensors and high-quality ground truth. Tasks supported by LuSNAR include 2D and 3D semantic segmentation, visual SLAM, LiDAR SLAM, stereo matching, and 3D reconstruction. Due to the controllable and editable features of lunar scenes in simulation engine, we divided the lunar surface simulation scene into 9 different levels based on the topographic relief level and surface object richness in order to cover different types of lunar surface. LuSNAR collected data under different terrains and landforms, which can be used as training samples to improve the generalization of the model. In the simulation scenes, the sensors carried by the simulated rover include a stereo camera, a LiDAR, and an IMU. Fig.~\ref{figure_1} shows nine scenes and visualized output data. Based on the simulation engine, reliable and high-precision ground truth are provided including depth, 2D semantic masks, 3D semantic labels, trajectory, and 3D point cloud of the scene, which are difficult to obtain using real sensors. To summarize, LuSNAR is a multi-task, multi-scene, and multi-sensor lunar surface dataset, which can effectively validate the feasibility of long-term autonomous environmental perception and navigation indicators.

The dataset have the following features:

\hangindent=2em
1) LuSNAR offered multiple tasks that consider the connections between segmentation, navigation and reconstruction, which is crucial for autonomous and safely driving for lunar rover. To validate the feasibility of the dataset, state-of-the-art 2D and 3D semantic segmentation, visual and LiDAR SLAM, and 3D reconstruction methods have been tested using the proposed dataset.

\hangindent=2em
2) LuSNAR dataset collected data in nine representative scenes to increase the diversity of dataset and explore the impact of topography steepness and object density on autonomous exploration tasks. Nine lunar simulation environments were designed based on different topographic relief and obejects density, and sorted in order according to the complexity of the terrain.

\hangindent=2em
3) The LuSNAR dataset contains multimodality ground truth labels synchronously generated for each trajectory, including depth, 2D semantic maps, 3D semantic labels, positions, and poses. The ground truth labels obtained through simulation engine have high reliability, providing strong support for ground verification and algorithm selection.

The structure of the paper is as follows. Section 2 presents a review of the current benchmark datasets for lunar and planetary scenes. Section 3 describes the characteristics and establishment process of LuSNAR dataset. The experimental results evaluating LuSNAR are discussed in Section 4. Finally, remarks and prospects for future research are discussed in Section 5.

\section{Related work}
In this section, public datasets, including extraterrestrial bodies, Mars and the Moon, for the tasks of segmentation and navigation of rovers, have been reviewed.
\subsection{Planetary Semantic segmentation datasets}
Roman et al.~\cite{Romain2019} presented an artificial lunar landscape dataset that simulates images of the moon using Terragen from Planetside Software. The Space Robotics Group at Keio University in Japan created this dataset, which is available on Kaggle and provides photorealistic images of the lunar surface along with semantic labels for training scene segmentation algorithms. It currently contains 9,766 realistic rendered images of the rocky lunar landscape and their corresponding semantic labels, including sky, smaller rocks, and larger rocks. It also includes bounding boxes of all larger rocks that can be used to train object detection algorithms.

Swan et al.~\cite{swan2021ai4mars} proposed the AI4Mars dataset, which consists of nearly 326K semantic segmentation full image labels on 35K images from the Curiosity, Opportunity, and Spirit rovers. The label data was collected through crowdsourcing, with an additional approximately 1,500 annotations from NASA mission's rover planners and scientists. The dataset includes four label types, namely soil, bedrock, sand, and large rock. It was built for training and validating terrain classification models for Mars. The current planetary semantic segmentation datasets primarily focus on the labels of 2D images which include fewer categories, lacking annotations for 3D data and failing to capture precise geometric information of scene objects. Consequently, they cannot contribute to the advancement of 3D semantic segmentation algorithms and overlook potential future research opportunities in exploring the fusion of 2D and 3D semantic information.

\subsection{Navigation datasets}
Due to the difficulties of acquiring real data from Mars and the Moon, most datasets are generated via field or software simulation. Given the geological and climatic similarities between Mars and Earth, it is feasible to find a location on Earth that is similar to the Martian environment. Tong et al.~\cite{tong2013canadian} presented a dataset collected at two distinct planetary analog rover test facilities in Canada, namely the University of Toronto Institute for Aerospace Studies (UTIAS) indoor rover test facility and the Canadian Space Agency's (CSA) Mars Emulation Terrain (MET). This dataset specifically focuses on 3D laser scans, with a total of 272 scans collected. Potential applications include terrain reconstruction, path planning, and LiDAR SLAM. Similarly, another dataset~\cite{lamarre2020canadian} was also collected by MET at CSA, but with different sensors. The sensor suite includes a color stereo camera, a monocular camera, an IMU, a pyranometer, drive power consumption monitors, wheel encoders, and a GPS receiver. The dataset contains 142,710 images from a stereo camera and 16,203 images from a monocular camera. It is divided into six separate runs, covering a total distance of over 1.2 km, and can be used for environment reconstruction, short-to-medium-distance path planning, omnidirectional visual-inertial odometry, and energy-aware planetary navigation.

The Erfoud dataset~\cite{lacroix2019erfoud} was acquired by two mobile robots, Mana and Minnie, at three different Mars-like locations in the Tafilalet region of Morocco. Collected along nine different trajectories totaling 13 km, the dataset contains approximately 110,000 georeferenced stereo image pairs and 40,000 LiDAR scans. It also includes wheel odometry data, FoG gyroscope measurements, IMU data, and pose ground truth obtained through RTK GPS. This dataset can be utilized for various applications including stereo vision, visual odometry, visual SLAM, terrain modeling, as well as more advanced tasks such as visual/LiDAR fusion, LiDAR SLAM, multi-robot SLAM, and absolute localization based on orbital data. The MADMAX dataset~\cite{meyer2021madmax} was collected in the same region as the Erfoud dataset. This dataset contains time-stamped recordings from a monochrome stereo camera, an omnidirectional stereo camera, a color camera, and an IMU. Additionally, it provides the 5 degrees of freedom (DoF) D-GNSS ground truth. There are 36 tracks in total, the longest track span is 1.5 km, and the total track length reaches 9.2 km. The dataset can be used as a benchmark for the accuracy and robustness of state-of-the-art navigation algorithms.

The Katwijk Beach Planetary Rover dataset~\cite{hewitt2018katwijk} was collected along a 1 km section of beach near Katwijk, the Netherlands. The beach was populated with a variety of artificial rocks in different sizes to emulate the conditions of Mars landing sites. The dataset can be divided into two parts. One part contains stereo images, pan and tilt orientations from a pan-tilt unit, scanning LiDAR and ToF measurements, IMU data, and RTK GPS positions. The other part contains the georeferenced images of UAV and DEMs. The dataset is well-suited for global and relative localization, SLAM, or related subtopics in environments where GNSS signals are unavailable.

Although the number of datasets about the Moon is smaller due to environmental differences, there are still some field-simulated lunar datasets. The Devon Island Rover Navigation Dataset~\cite{furgale2012devon} was collected at a Mars/Moon analog site on Devon Island, Nunavut. Mainly includes stereo images, 3D laser ranging scans, positions from the DGPS, sun vectors, inclinometer data, etc. It can be used for studying localization problems in GPS-denied environments. Mount Etna in Sicily, Italy, is also an excellent environment similar to the Moon and Mars. Both the LRNT dataset~\cite{vayugundla2018datasets} and the S3LI dataset~\cite{giubilato2022challenges} capture data at this site. The former was captured by the Lightweight Rover Unit (LRU), which traversed a distance of approximately 1 km. It provides grayscale images, dense depth images, IMU data, wheel odometry estimates, and ground truth from DGPS. This dataset complements the MADMAX dataset by recording simulated data of the Moon, using the same sensor settings. It evaluated the existing pose estimation system running on the rover and proved its importance to fuse visual and inertial navigation systems in GPS-denied and unstructured planetary environments. The latter captured 7 sequences ranging from 8 to 30 minutes, covering a distance of up to 1.3 km, collected by a handheld sensor suite that includes a monochrome stereo camera, a solid-state LiDAR, an IMU, and a GPS receiver. The dataset is used for the evaluation of diverse visual-inertial SLAM algorithms, emphasizing their advantages and limitations, while also presenting examples of potential use cases. 

The existing navigation datasets are limited in terms of data types and lack support for simultaneous multitasking, which hinders the system-level integration of multiple modules in a loosely coupled manner to enhance overall performance. Most of these datasets focus on Mars, with only a small number of terrain scenes available. This lack of environmental diversity restrains the improvement of rovers' navigation capabilities in dynamic and unknown environments. Furthermore, they fail to provide true depth information of the scene and some datasets lack complete pose ground truth, making effective evaluation of 3D reconstruction and localization tasks difficult.

Table~\ref{table_1} shows a comparison between the state-of-the-art datasets and the LuSNAR dataset proposed in this paper. It can be seen that none of the previous works can simultaneously support segmentation, navigation and reconstruction tasks in one dataset, while the LuSNAR dataset can do this. The LuSNAR dataset comprises various data types from stereo camera, LiDAR, and IMU, offering high-precision 2D/3D semantic labels, pose ground truth, and depth maps generated by the simulation engine. This facilitates multi-task testing and evaluation of algorithms for 2D/3D semantic segmentation, localization, and 3D reconstruction. Moreover, this dataset collects data from 9 lunar surface scenes featuring unique terrains, which not only exceeds other datasets in the number of scenes and solves the scarcity issue of lunar surface datasets, but also helps rovers to enhance the generalization ability in unknown lunar surface scenes through the diversity of environments.

\begin{table*}[!t]
\centering
\caption{Comparison of the state-of-the-art datasets with the proposed LuSNAR dataset.}
\label{table_1}
\renewcommand{\arraystretch}{1.3}
\setlength{\tabcolsep}{4pt}
\begin{tabular}{|C{2.0cm}|C{1.6cm}|C{1.6cm}|C{0.9cm}|C{0.7cm}|C{0.7cm}|C{0.8cm}|C{0.7cm}|C{0.7cm}|C{0.9cm}|C{0.7cm}|C{0.8cm}|C{0.8cm}|}
\hline
\multirow{2}{*}{\raisebox{-2.0\height}{Dataset}} & \multirow{2}{*}{\raisebox{-2.0\height}{Scene}} & \multirow{2}{*}{\raisebox{-2.0\height}{Real/Synthetic}} & \multirow{2}{*}{\raisebox{-1.0\height}{\shortstack{Scenes\\Number}}} & \multicolumn{4}{c|}{Sensors} & \multicolumn{5}{c|}{Ground Truth} \\
\cline{5-13}
 &  &  &  & \raisebox{-0.9\height}{Mono} & \raisebox{-0.9\height}{Stereo} & \raisebox{-0.9\height}{LiDAR} & \raisebox{-0.9\height}{IMU} & \raisebox{-0.9\height}{Depth} & \raisebox{-0.9\height}{Position} & \raisebox{-0.9\height}{Pose} & \multicolumn{2}{c|}{Semantic-Label} \\
\cline{12-13}
 &  &  &  &  &  &  &  &  &  &  & 2D & 3D \\
\hline
Lunar Landscape & Moon & Synthetic & 1 & \cmark & \xmark & \xmark & \xmark & \xmark & \xmark & \xmark & \cmark & \xmark \\
\hline
AI4MARS & Mars & Real & 3 & \cmark & \cmark & \xmark & \xmark & \xmark & \xmark & \xmark & \cmark & \xmark \\
\hline
Canadian 3D Mapping & Mars & Real & 2 & \xmark & \xmark & \cmark & \cmark & \xmark & \cmark & \cmark & \xmark & \xmark \\
\hline
Canadian Energy Aware & Mars & Real & 1 & \cmark & \cmark & \xmark & \cmark & \xmark & \cmark & \xmark & \xmark & \xmark \\
\hline
Erfoud & Mars & Real & 3 & \cmark & \cmark & \cmark & \cmark & \xmark & \cmark & \cmark & \xmark & \xmark \\
\hline
MADMAX & Mars & Real & 8 & \cmark & \cmark & \xmark & \cmark & \xmark & \cmark & \cmark & \xmark & \xmark \\
\hline
Katwijk Beach & Mars & Real & 1 & \cmark & \cmark & \cmark & \cmark & \xmark & \cmark & \xmark & \xmark & \xmark \\
\hline
Devon Island & Mars/Moon & Real & 4 & \cmark & \cmark & \cmark & \xmark & \xmark & \cmark & \xmark & \xmark & \xmark \\
\hline
LRNT & Mars/Moon & Real & 1 & \cmark & \cmark & \xmark & \cmark & \xmark & \cmark & \xmark & \xmark & \xmark \\
\hline
S3LI & Mars/Moon & Real & 7 & \cmark & \cmark & \cmark & \cmark & \xmark & \cmark & \xmark & \xmark & \xmark \\
\hline
Ours & Moon & Synthetic & 9 & \cmark & \cmark & \cmark & \cmark & \cmark & \cmark & \cmark & \cmark & \cmark \\
\hline
\end{tabular}
\end{table*}

\section{Dataset features}
The LuSNAR dataset is based on simulation engine to generate a multi-task, multi-scene, and multi-label lunar surface dataset, which can be used for ground verification, algorithm selection of autonomous environmental perception, and navigation of lunar rovers. To achieve this, diverse and realistic lunar scenes are designed for data collection, aiming to equip rovers with the ability to generalize when encountering unknown environments. In this section, a detailed overview of the principles behind the simulation scene design, as well as the content and features of the LuSNAR dataset is provided.

\subsection{Multi-task supported}
\subsubsection{Semantic perception}
The semantic information has important practical significance in lunar surface exploration missions. It can not only provide obstacle information to help rovers assess terrain traversability, but also provide prior knowledge for lunar geological research, allowing scientists to select landforms of interest for in-depth investigation~\cite{garcia2020review}. Both the image sequence and point cloud sequence in the LuSNAR dataset contain semantic labels, enabling the evaluation of 2D and 3D semantic segmentation algorithms for lunar scenes. The data taken from cameras and LiDAR have their merits and problems. The camera can provide rich information of appearance, texture, and color of lunar surface; however, it is susceptible to illumination variations and unable to detect obstacles in shadows, thereby features extracted from images may fail to be tracked during travel. On the other hand, LiDAR can offer precise three-dimensional coordinates of objects with geometric information and remains insensitive to illumination changes. Nevertheless, the point cloud obtained from scanning is usually relatively sparse and lacks comparable levels of detail as images. Therefore, combining camera and LiDAR data can use the privilege of both sensor and provide more reliable result for semantic segmentation. The LuSNAR dataset can verify the applicability and complementarity of 2D and 3D semantic perception algorithms in lunar surface scenarios, which contributes to the high-level visual understanding research on lunar surface.

\subsubsection{SLAM}
Autonomous navigation tasks are a crucial technology for lunar rovers in autonomous lunar exploration, including key steps such as mapping, localization, and path planning. SLAM (Simultaneous Localization and Mapping) technology can calculate the localization, velocity, and orientation of the rover based on sensors carried by the rover itself, without relying on ground support, supporting the lunar rover's autonomous exploration mission. The LuSNAR dataset supports multiple sensor SLAM solutions, including monocular SLAM, stereo SLAM, LiDAR SLAM, and IMU (Inertial Measurement Unit) fusion-based SLAM. Visual SLAM has the advantages of low cost and rich information of cameras. However, visual sensors are highly sensitive to changes in lighting conditions, which can affect localization accuracy. LiDAR SLAM can directly acquire three-dimensional information about the environment, achieving higher precision pose estimation and mapping by registering point cloud data from adjacent frames. However, LiDAR SLAM may degrade or fail in scenes without obvious geometric features. Multi-sensor SLAM solutions that incorporate IMU data can address challenges related to global localization in similar geometric environments and environmental changes~\cite{xu2022review}. This helps to correct errors in cases where visual and LiDAR information is temporarily missing, thus improving the robustness of autonomous navigation systems. The LuSNAR dataset provides multimodal data for lunar rover navigation and mapping on the lunar surface, including camera images, LiDAR scan sequences, IMU data, and ground truth pose information. This facilitates the validation of SLAM algorithms based on different sensors for localization accuracy and real-time performance, further promoting the application of SLAM technology in lunar rover autonomous exploration missions.

\subsubsection{3D Reconstruction}
The lunar surface is an uneven, irregular, and unstructured environment with various obstacles. To ensure the safe navigation of lunar rovers during their missions, it is essential to perform dense terrain reconstruction of the surroundings for path planning, target area identification and obstacle avoidance. A stereo-matching algorithm based on binocular vision is an effective method for dense terrain reconstruction~\cite{laga2020survey}. This algorithm calculates a disparity map from the stereo images and recovers a 3D point cloud of the scene using camera parameters. The LuSNAR dataset provides dense depth ground truth and camera parameters for the lunar rover's surrounding environment, allowing for the evaluation of the accuracy and real-time performance of stereo-matching algorithms in terrain reconstruction. In practical applications, relying only on point clouds obtained from individual camera stations provides limited information about the lunar rover's surroundings. It is necessary to merge and fuse terrain reconstruction results from multiple camera stations by combining their poses. The LuSNAR dataset also synchronously provides ground truth poses of the lunar rover, enabling the fusion and stitching of multi-station point cloud data to generate point cloud of large area on lunar surface. Dense terrain reconstruction data accurately reflect changes in the surrounding terrain, serving as fundamental information for lunar rover localization, path planning, scientific exploration, and the overall goal of achieving a comprehensive perception of the rover's environment.

\subsection{Diversity in lunar topographic relief and objects density}
Lunar simulation scenes are designed and rendered based on Unreal Engine 4, utilizing the AirSim plugin developed by Microsoft for data collection~\cite{shah2018airsim}. UE4 is a game development engine created by Epic Games, and we leveraged lunar-related terrain, features, and material textures available in the Unreal Marketplace as our assets. During the simulation scene design, we referenced Apollo lunar exploration data and images captured by China's Yutu-2 lunar rover to ensure the accuracy and scientific fidelity of the scenes. According to the geological characteristics of the actual lunar surface, the features in the scene are divided into five types: lunar regolith, rocks, impact craters, mountains and sky.

\begin{table}[h]
\centering
\caption{The Comparison of rock size between the landing areas of seven actual lunar exploration missions and the simulated scenes of the LuSNAR dataset.}
\label{table_2}
\begin{tabular}{|c|c|c|c|}
\hline
Landing area & \makecell{Average\\resolution\\(m/pixel)} & \makecell{Maximum\\diameter\\(m)} & \makecell{Minimum\\Diameter\\(m)} \\
\hline
CE-3         & 1.51               & 19.0             & 1.51             \\
\cline{1-4}
Apollo11     & 0.75               & 3.63             & 0.94             \\
\cline{1-4}
Apollo16     & 0.47               & 2.72             & 0.48             \\
\cline{1-4}
Surveyor I   & 0.81               & 3.98             & 0.76             \\
\cline{1-4}
Surveyor III & 0.43               & 2.87             & 0.47             \\
\cline{1-4}
Surveyor VI  & 0.66               & 4.00             & 0.50             \\
\cline{1-4}
Surveyor VII & 0.51               & 5.27             & 0.45             \\
\cline{1-4}
LuSNAR       & -                  & 5.45             & 0.13             \\
\hline
\end{tabular}
\end{table}

Table~\ref{table_2}  presents the rock size statistics for the landing areas of seven actual lunar exploration missions and the simulated scenes of the LuSNAR dataset. Each mission utilized 2-3 high-resolution LRO NAC images to identify the rock sizes in the lunar landing areas~\cite{li2017rock}.

The size of the impact crater in the simulation scenario of the LuSNAR dataset was designed with reference to the impact crater in the CE-5 landing zone.

According to the statistical results of 231 meteorite craters in the CE-5 landing zone, the diameter range of impact craters is from 7.0m to 371.2m, and 91.0\% of the craters have a diameter less than 100m. 77.1\% of the craters have a depth greater than 1m, and there are 7 craters with a depth of more than 10m. The d/D ratio of meteorite craters in the CE-5 landing zone ranges from 0.018 to 0.134, with an average value of 0.055~\cite{bo2022catalogue}.

\begin{table}[h]
\centering
\caption{The diameter and depth statistics of impact craters in 9 scenes.}
\label{table_3}
\begin{tabular}{|c|c|c|c|}
\hline
Scene  & Diameter (m) & Depth (m) & d/D ratio \\
\hline
\multirow{1}{*}{Scene 2}  & 69.50  & 5.37  & 0.07 \\
\hline
\multirow{1}{*}{Scene 3}  & 73.31  & 6.07  & 0.08 \\
\hline
\multirow{3}{*}{Scene 4}  & 49.10  & 6.78  & 0.13 \\
         & 25.61  & 3.69  & 0.14 \\
         & 62.00  & 6.85  & 0.11 \\
\hline
\multirow{3}{*}{Scene 5}  & 63.63  & 11.48 & 0.18 \\
         & 35.05  & 3.73  & 0.10 \\
         & 65.25  & 7.13  & 0.10 \\
\hline
\multirow{3}{*}{Scene 6}  & 27.47  & 5.30  & 0.19 \\
         & 24.22  & 2.41  & 0.09 \\
         & 67.31  & 11.56 & 0.17 \\
\hline
\multirow{5}{*}{Scene 7}  & 24.64  & 2.53  & 0.10 \\
         & 67.77  & 11.11 & 0.16 \\
         & 20.82  & 2.00  & 0.09 \\
         & 14.54  & 1.69  & 0.11 \\
         & 33.72  & 3.41  & 0.10 \\
\hline
\multirow{4}{*}{Scene 8}  & 52.36  & 8.04  & 0.15 \\
         & 27.91  & 2.82  & 0.10 \\
         & 100.68 & 6.56  & 0.06 \\
         & 59.35  & 8.00  & 0.13 \\
\hline
\multirow{4}{*}{Scene 9}  & 22.88  & 3.57  & 0.15 \\
         & 59.64  & 5.11  & 0.08 \\
         & 22.11  & 1.16  & 0.05 \\
         & 32.19  & 1.73  & 0.05 \\
\hline
\end{tabular}
\end{table}

As shown in the Table~\ref{table_3}, the diameter and depth statistics of all impact craters included in the simulation scene are shown. The diameter and depth of impact craters in the Lusnar dataset scene are consistent with the size of impact craters in real lunar scenes.

In the LRO NAC images, due to limitations in image resolution, some clusters of small rocks cannot be accurately identified. Consequently, the minimum diameter of rocks recorded in the table is constrained by the resolution. The LuSNAR dataset simulates based on the rock sizes in the landing areas of actual lunar exploration missions and incorporates rocks with diameters ranging from 0.1 to 0.3 meters to enrich the diversity of the scenarios. The impact craters in the LuSNAR dataset are simulated with reference to the area of the Apollo 16 lunar exploration mission, where the diameter range of impact craters is approximately 10 meters to 100 meters. The diameter range of impact craters in the nine scenarios of the LuSNAR dataset aligns with this range.

\begin{figure*}[!t]
    \centering
    \subfloat[\label{figure_2_1}]{
        \includegraphics[width=0.50\textwidth]{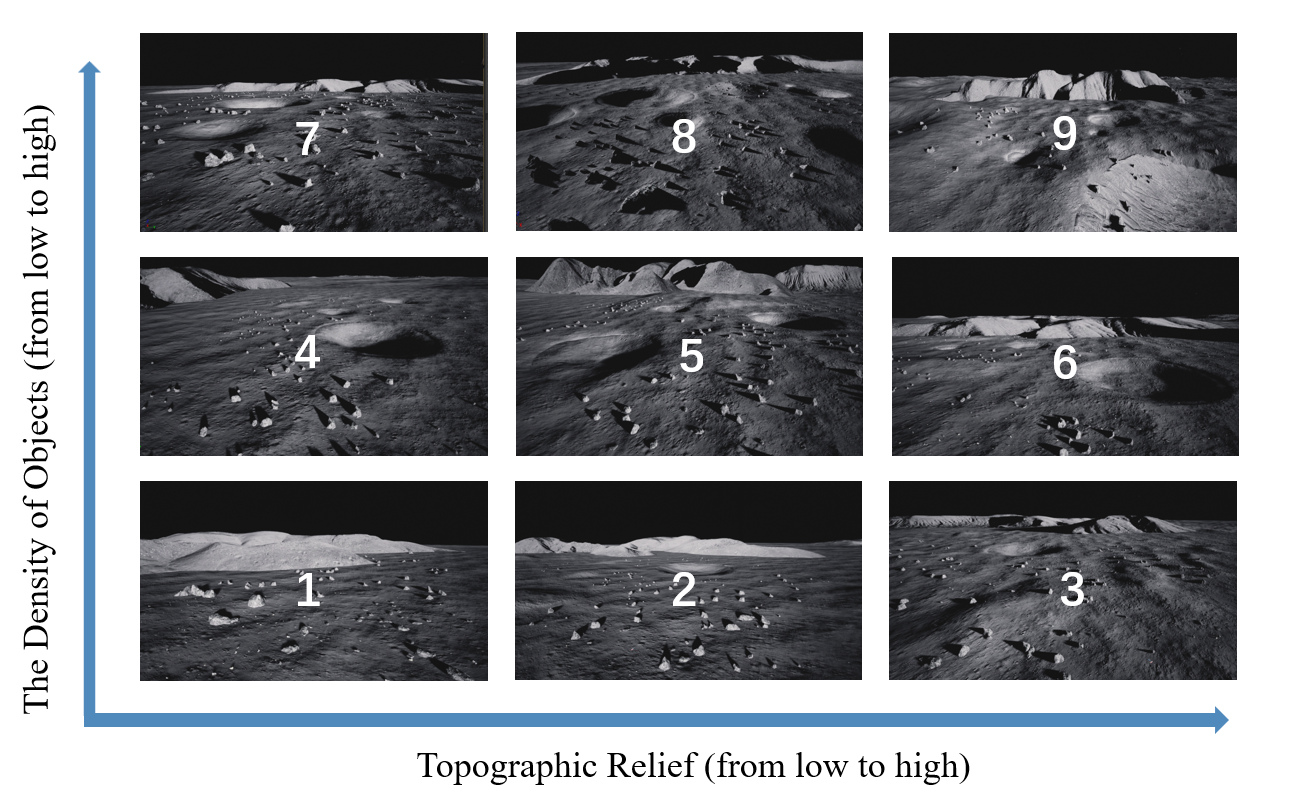}
    }
    \hfill
    \subfloat[\label{figure_2_2}]{
        \raisebox{0.5cm}{
            \includegraphics[width=0.45\textwidth]{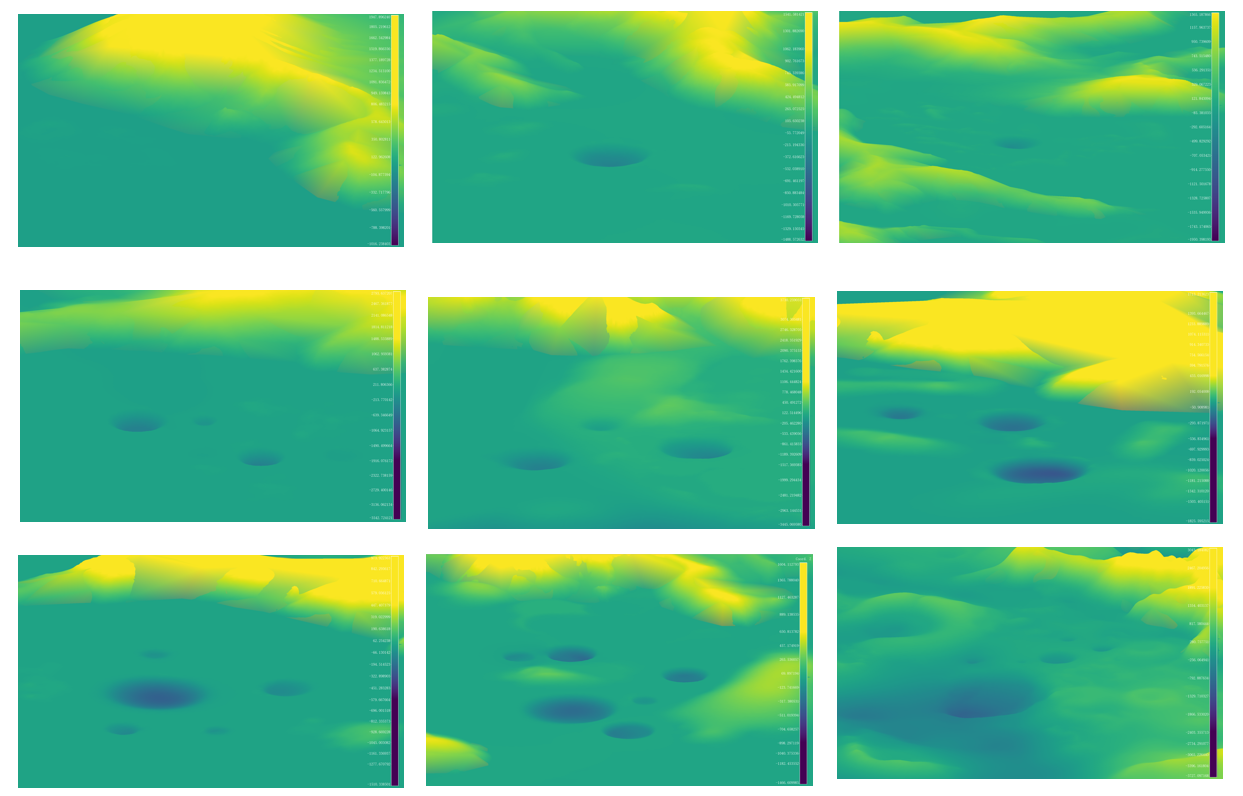}
        }
    }
    
    \vspace{0.5cm}
    
    \hspace{1.0cm}
    \subfloat[\label{figure_2_3}]{
        \includegraphics[width=0.35\textwidth]{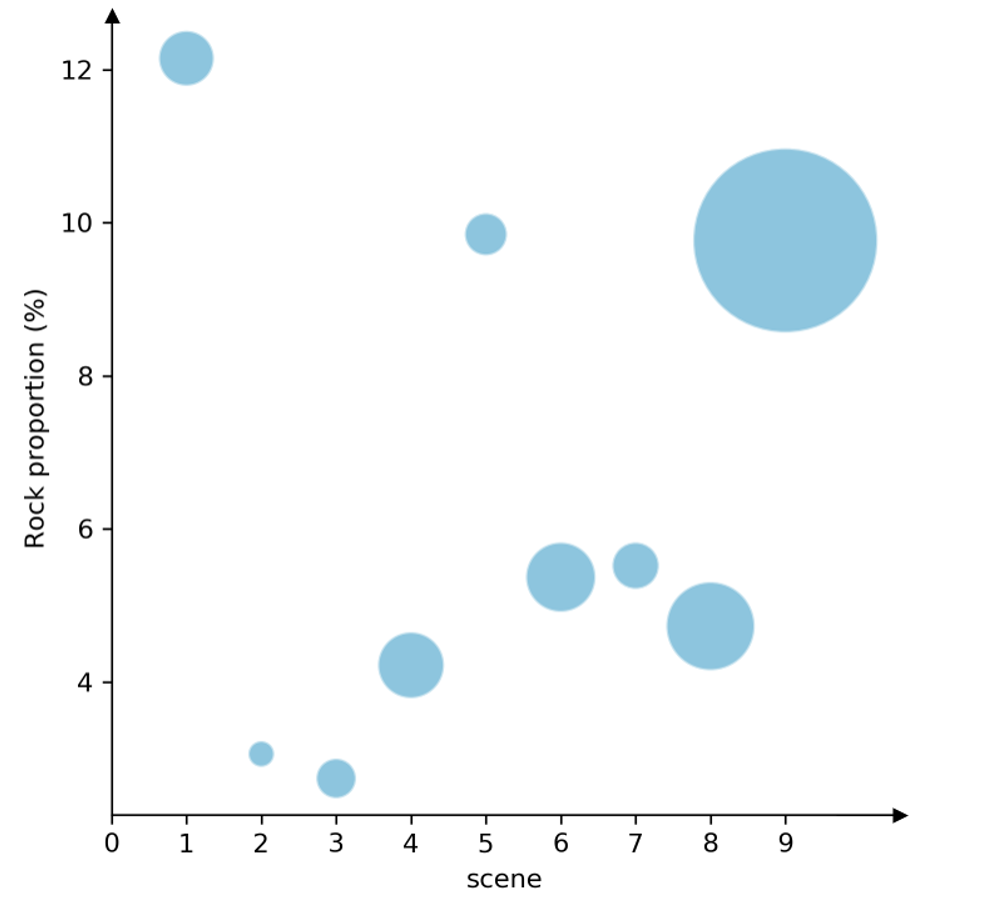}
    }
    \hfill
    \subfloat[\label{figure_2_4}]{
        \includegraphics[width=0.45\textwidth]{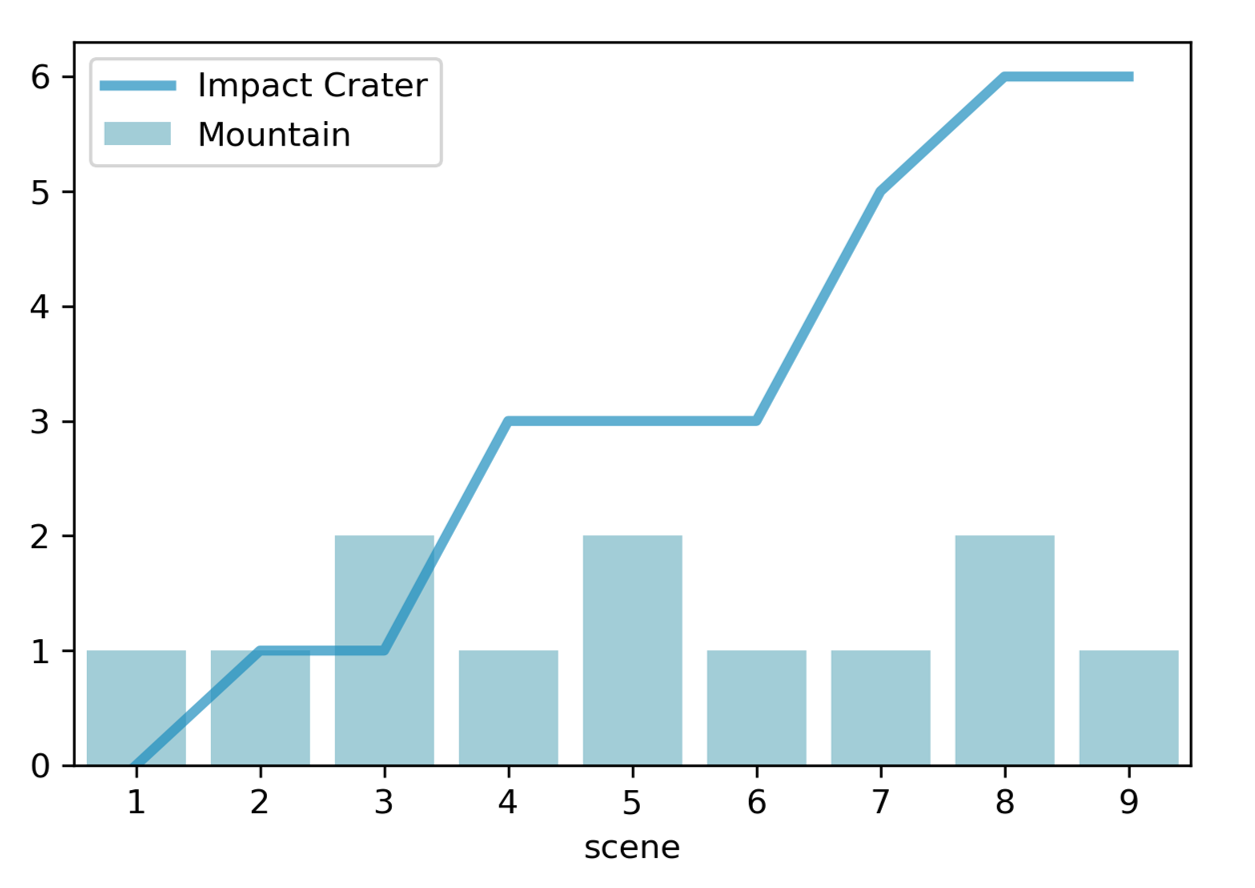}
    }
    
    \caption{Qualitative and quantitative comparison of the 9 scenes. (a) The topographic relief and density of objects in 9 scenes. (b) The topographic relief trends of the 9 scenes. (c) The abundance of rocks in 9 scenes. (d) The statistics on the number of mountain ranges and impact craters in 9 scenes.}
    \label{figure_2}
\end{figure*}

The topographic relief and obejects density of the lunar exploration areas can be largely different due to scientific purpose, leading to distinct planning and requirements for autonomous exploration tasks. To ensure the diversity lunar terrain topography and features of the dataset, this paper innovatively used topographic relief and the density of objects as two dimensions and designed nine lunar surface scenes with discrimination. As shown in Fig.~\ref{figure_2}\subref{figure_2_1}, the horizontal axis represents topographic relief from gentle to steep, while the vertical axis represents feature richness from sparse to abundant. The two dimensions each have three different levels, and nine scenarios simulate a variety of lunar environments in the form of levels. This approach not only enriches scene diversity but also facilitates the exploration of how diverse terrains and topographies influence environmental perception and autonomous navigation algorithms. Fig.~\ref{figure_2}\subref{figure_2_2} illustrates the topographic relief trends of the nine scenes and renders them with different colors based on elevation information. Scenes 1, 4, and 6 exhibit relatively flat terrain, while scenes 2, 5, and 7 feature minor undulations, and scenes 3, 6, and 9 display significant terrain variations. Additionally, this paper proposed a quantitative analysis of the types of features in these nine scenes. The quantity of rocks and impact craters served as the primary criteria for assessing feature richness in the scenes. Fig.~\ref{figure_2}\subref{figure_2_3} presents the abundance of rocks in the lunar rover's travel area within the nine scenes, with the vertical axis indicating the proportion of rocks in the lunar rover's path and the circular area representing the total quantity of rocks in the scene. Fig.~\ref{figure_2}\subref{figure_2_4} provides statistics on the number of mountain ranges and impact craters in the nine scenes, indicating an overall increasing trend in the number of impact craters. Due to the larger volume of mountains and their limited occurrence within the field of view, the number of mountains in each scene is limited to 1-2. Considering the distribution of rocks and the number of impact craters, scenes 7, 8, and 9 emerge as the three scenes with the richest features. In summary, the terrain complexity gradually increases from scenes 1 to 9, providing lunar surface scene data with various combinations of terrain and topography.

\subsection{multi-label}
\begin{figure*}[!t]
\centering
\includegraphics[width=\textwidth]{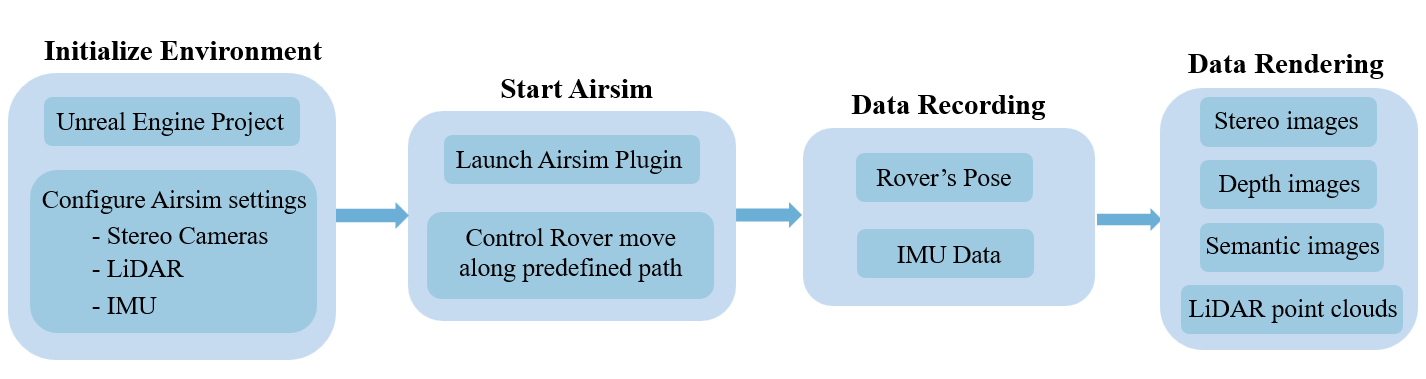}
\caption{Simulation data generation pipeline.}
\label{figure_3}
\end{figure*}

This paper uses the "AirSimGameMode" mode for simulation in Unreal Engine 4. The simulation involves manually controlling a simulated rover to navigate within the scene. Attention has been given to ensuring diversity in both the scene and motion modes to ensure its functionality in various terrains and environments. The data generation process based on Unreal Engine 4 and the Airsim plugin is shown in the Fig.~\ref{figure_3}. Firstly, the environment is initialized, and sensors in Airsim are configured within the constructed lunar scene, setting the parameters for stereo cameras, LiDAR, and IMU respectively.  Then, UE and Airsim are launched, and the Rover is controlled to move along a predefined path. The Airsim API is called to online record the Rover's ground truth poses and IMU data.  Finally, using the recorded poses and sensor extrinsics, stereo images, depth images, and semantic images are offline rendered, and LiDAR point clouds are generated. All data is then stored for further use.

\begin{table*}[!t]
    \centering
    \caption{Parameters of stereo cameras, LiDAR and IMU in the simulation engine.}
    \label{table_4}
    \begin{tabular}{|c|p{12cm}|}
        \hline
        Sensor & Details \\ \hline
        \raisebox{-0.5\height}{2$\times$Camera} & RGB, 10 Hz capture frequency, 1024$\times$1024 resolution, 610.17784 focal length, 80$^\circ$$\times$80$^\circ$ FOV, 310 mm baseline. \\ \hline
        \raisebox{-0.5\height}{1$\times$LiDAR} & Spinning, 128 beams, 10Hz capture frequency, 360$^\circ$ horizontal FOV, -25$^\circ$ to 27$^\circ$ vertical FOV, $\leq$30 m range, up to 20M points per second. \\ \hline
        \raisebox{-1.5\height}{1$\times$IMU} & 100 Hz capture frequency, 3-Axis, 0.002353596(m/s$^3$$\sqrt{\text{Hz}}$) accelerometer random walk, 8.7266462e-5(rad/s$\sqrt{\text{Hz}}$) gyroscope random walk, 1.2481827e-5(m/s$^2$$\sqrt{\text{Hz}}$) accelerometer bias instability, 9.9735023e-7(rad/s$\sqrt{\text{Hz}}$) gyroscope bias instability. \\ \hline
    \end{tabular}
\end{table*}

The rover collects trajectories from each of the nine scenes, and its sensors move along with the trajectory to gather multimodal data. The rover's motion modes include straight-line movement, turning, climbing, descending, obstacle avoidance, and circumnavigation, all of which are representative of scenarios encountered in real-world tasks. The coverage of each scene is about 300 meters × 300 meters. The rover's travel distance ranges from 200m to 600m, and its speed during operation does not exceed 5m/s. The simulation sensor parameters and their corresponding output data are shown in Table~\ref{table_4}. It should be noted that it is difficult for simulated sensors to accurately copy all the characteristics of real sensors. For example, camera distortion and installation matrix error are not considered in the sensor simulation of binocular camera in this paper. Radar and IMU may encounter noise, dynamic range limitation and other problems under lunar surface conditions, which may not be completely and accurately reproduced in the simulation. The LuSNAR dataset comprises a collection of 13,006 sequences, gathered from nine scenes, each with a consistent set of internal and external parameters. The sequence length of each scene is about 1000 to 2000. Each sequence includes stereo image pairs, single-frame point clouds, semantic labels, IMU data, and rover pose ground truth. Each sequence contains the following data:

\begin{itemize}
    \item RGB image @ 1024 x 1024 in PNG.
    \item 2D semantic segmentation label @ 1024 x 1024 in PNG. For each pixel of the left-sided image, an ID of the represented object is encoded.
    \item Depth map @ 1024 x 1024 in 16-bit PFM. For each pixel of the left-sided image, the true depth of the represented object is stored.
    \item 3D semantic point cloud on the order of $10^4$ in TXT. Each point's ID represents a different type of object.
    \item IMU data in TXT.
    \item 3D ground truth poses in TXT.
    \item Timestamps for the cameras, LiDAR, and IMU.
\end{itemize}

All data is named according to timestamps, with images and semantic labels stored in PNG format, depth maps in PFM format, and IMU, ground truth pose, and 3D LiDAR point cloud data stored in TXT format.

\begin{figure*}[!t]
    \centering
    \subfloat[\label{figure_4_1}]{
        \includegraphics[width=0.44\textwidth]{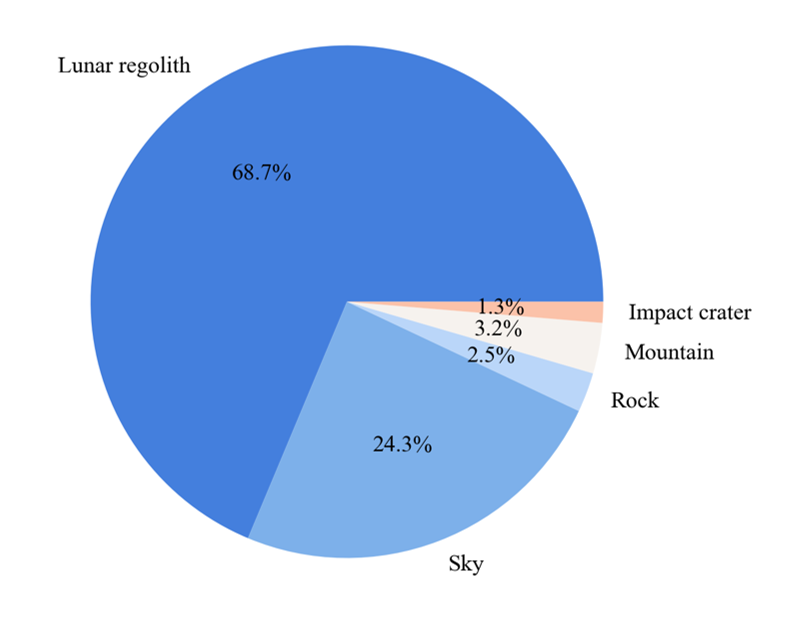}
    }
    \hfill
    \subfloat[\label{figure_4_2}]{\raisebox{0.3cm}{
            \includegraphics[width=0.50\textwidth]{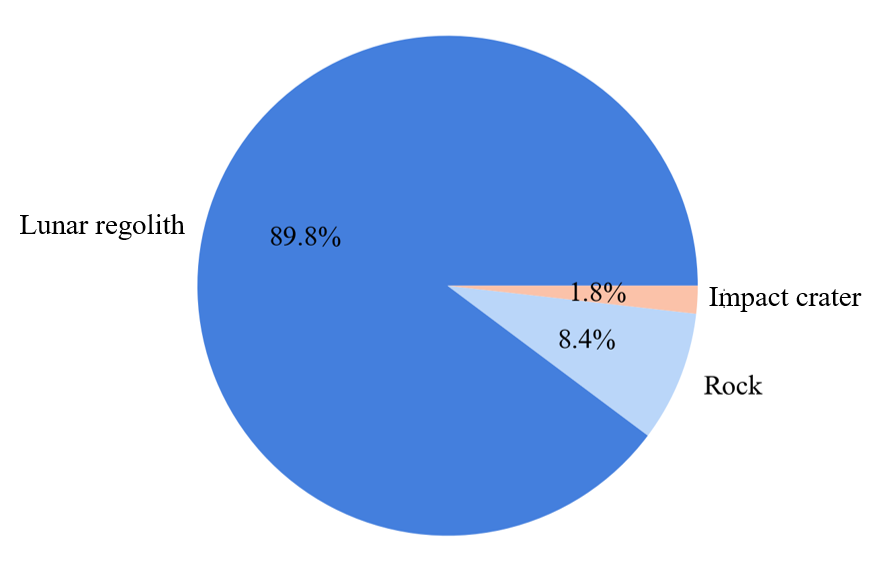}}
    }
    \vspace{0.5cm}   
    \hspace{1.0cm}
    \subfloat[\label{figure_4_3}]{
        \includegraphics[width=0.45\textwidth]{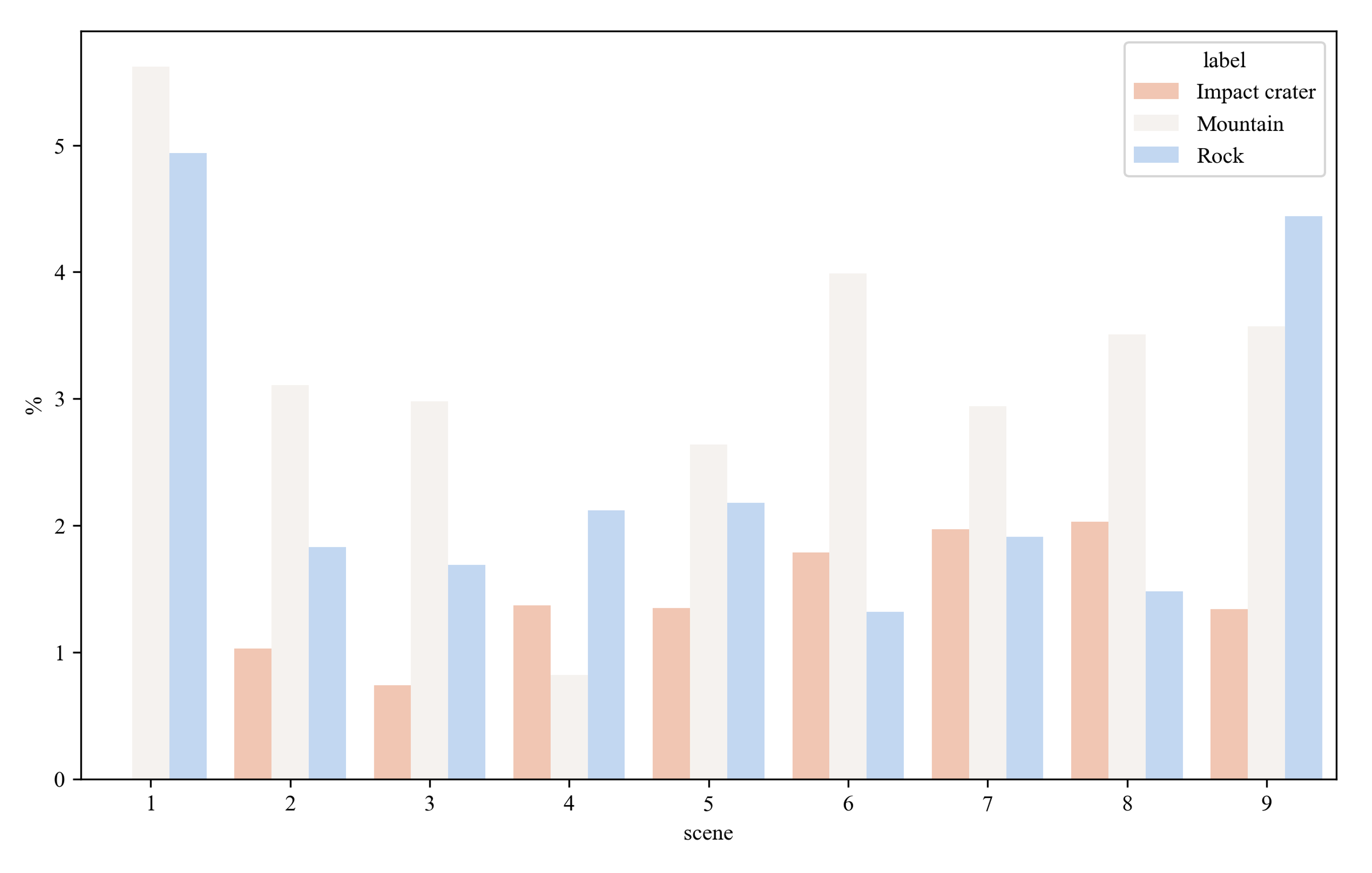}
    }
    \hfill
    \subfloat[\label{figure_4_4}]{
        \includegraphics[width=0.45\textwidth]{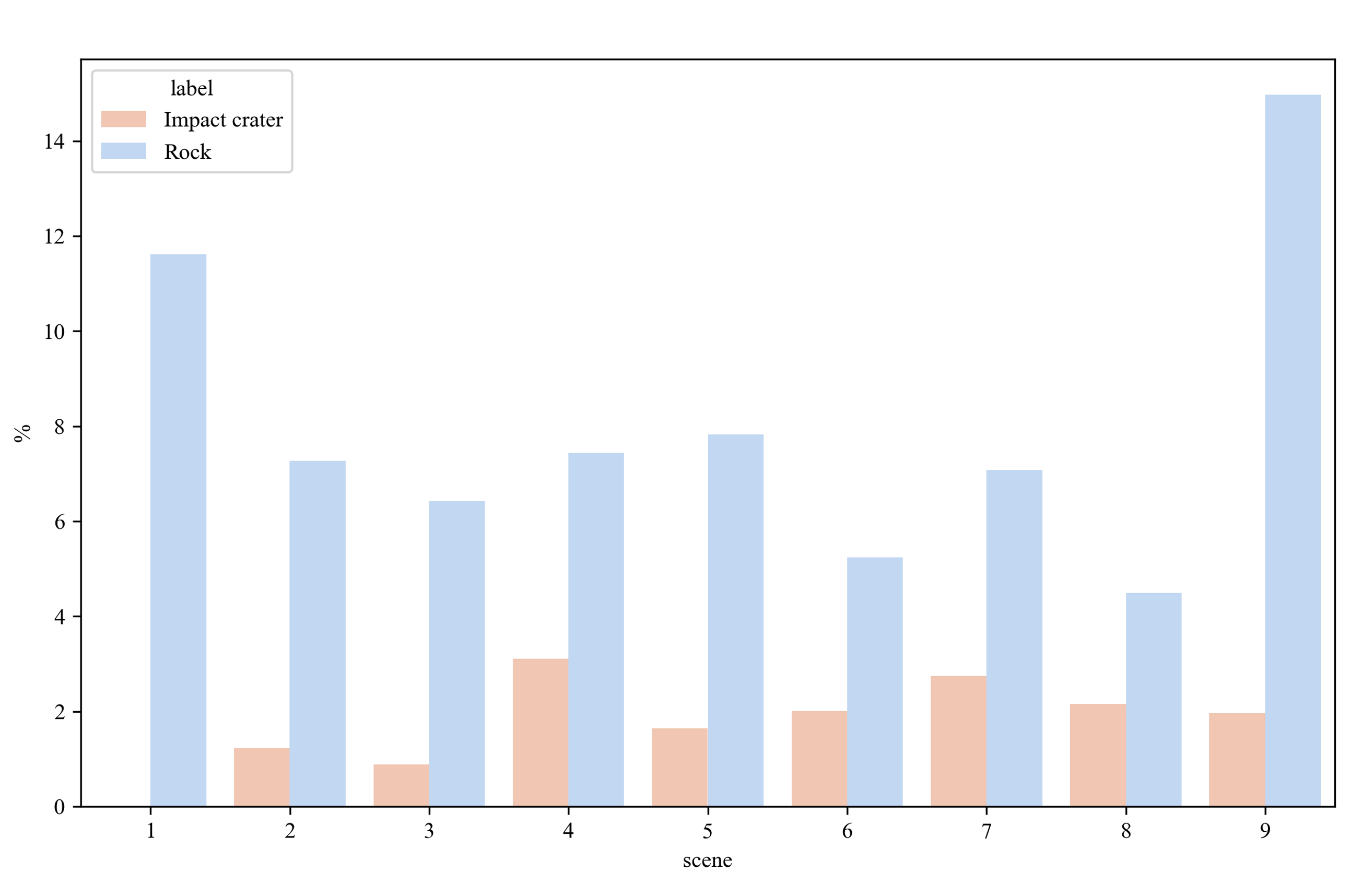}
    }
    \caption{The statistical distribution of semantic labels in the LuSNAR dataset. (a) The proportion of different object quantities in images across all scenes. (b) The proportion of different object quantities in LiDAR point clouds across all scenes. (c) The proportion of object quantities in images across 9 different scenes. (d) The proportion of object quantities in LiDAR point clouds across 9 different scenes.}
    \label{figure_4}
\end{figure*}

Fig.~\ref{figure_4} shows the statistical distribution of semantic labels in the LuSNAR dataset. Fig.~\ref{figure_4}\subref{figure_4_1} and Fig.~\ref{figure_4}\subref{figure_4_2} respectively illustrate the proportion of different object quantities in images and LiDAR point clouds across all scenes, while Fig.~\ref{figure_4}\subref{figure_4_3} and Fig.~\ref{figure_4}\subref{figure_4_4} respectively compare the proportion of object quantities in images and LiDAR point clouds across 9 different scenes.

\begin{figure}[h]
\centering
\includegraphics[width=3.6in]{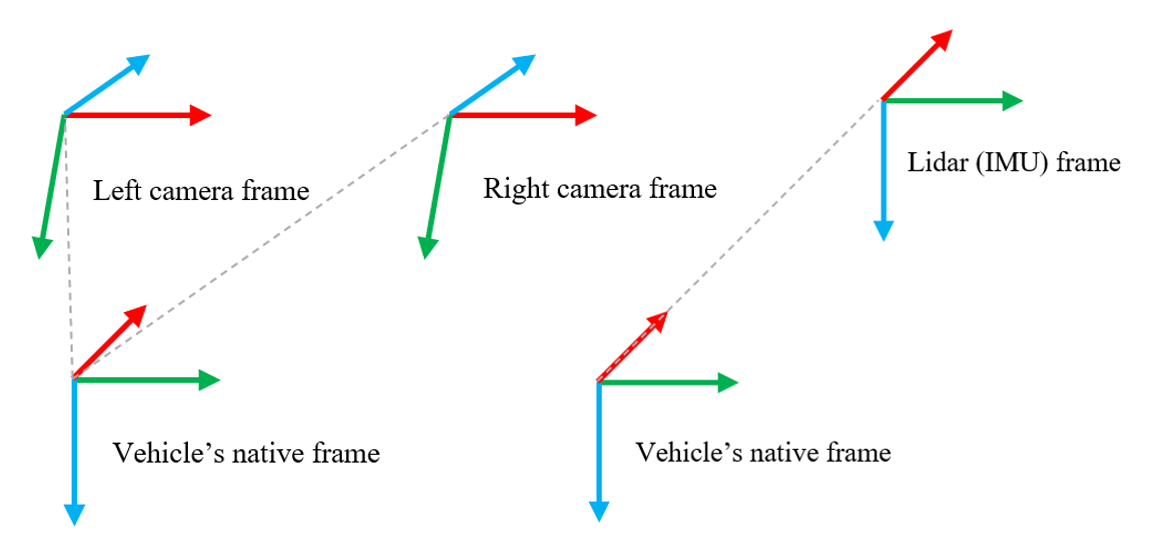}
\caption{Extrinsic settings for rover and sensors.}
\label{figure_5}
\end{figure}

The lunar rover’s native frame and sensor frame defined in the simulation engine are shown in Fig.~\ref{figure_5}. The forward direction of the lunar rover is the positive X-axis, the horizontal direction to the right is the positive Y-axis, and the vertical downward direction is the positive Z-axis. Taking the center of the rover as frame origin, both LiDAR and IMU are positioned at (1, 0, -1.5), aligning their frame orientations with that of the lunar rover. Regarding navigation camera, the forward direction is the positive Z-axis, the horizontal right direction is the positive X-axis, and the vertical downward direction is the positive Y-axis. Additionally, the navigation camera has a pitch angle of -20 degrees. The left navigation camera is positioned at (1, -0.155, -1.5) with the center of the lunar rover as the frame origin, while the right navigation camera is positioned at (1, 0.155, -1.5).

\begin{figure*}[!t]
    \centering
    \subfloat[\label{figure_6_1}]{
        \includegraphics[width=\textwidth]{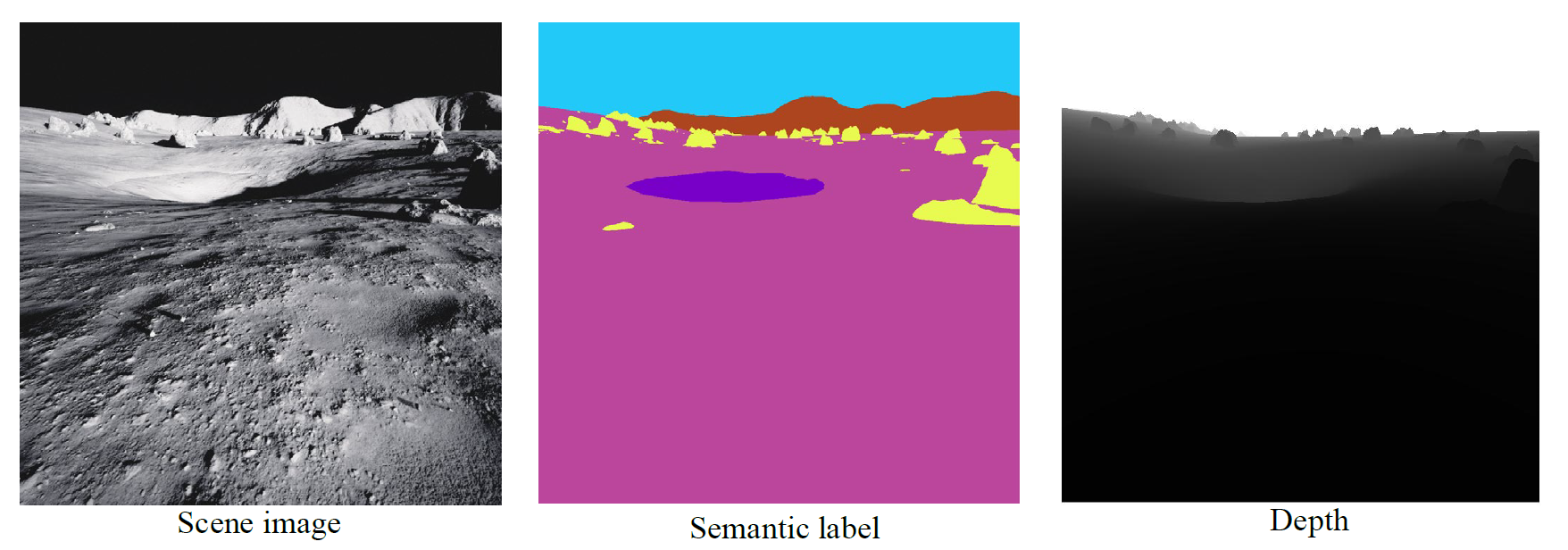}
    }
    \hfill
    \subfloat[\label{figure_6_2}]{
            \includegraphics[width=\textwidth]{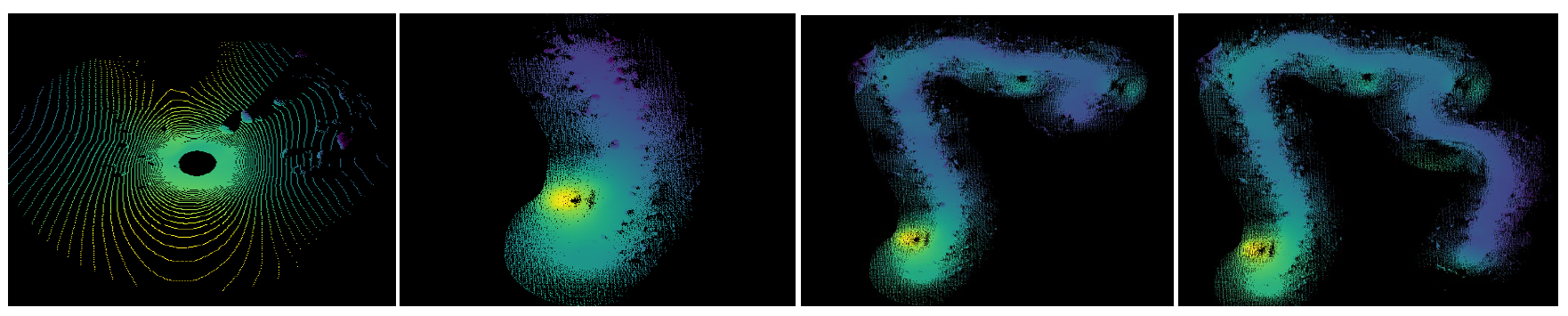}
    }
    \caption{An example of 2D labels and 3D trajectories. (a) 2D semantic image and depth map. (b) Ground truth point cloud map construction process.}
    \label{figure_6}
\end{figure*}

An example of labels generated in the proposed dataset are shown in Fig.~\ref{figure_6}\subref{figure_6_1}, which demonstrate the collected stereo image pairs, depth maps, and semantic labels. The LiDAR point cloud frame is sequentially loaded to obtain trajectories in different time periods, as shown in Fig.~\ref{figure_6}\subref{figure_6_2}. In the 2D image semantic segmentation mask, object labels include lunar regolith, lunar rocks, impact craters, mountains, and the sky. In the 3D point cloud semantic segmentation mask, object labels include lunar regolith, rocks, and impact craters. The LuSNAR dataset has a total size of 108GB and provides more than 42GB of stereo image pairs, 50GB of depth maps, 356MB of semantic segmentation labels, and 14GB of single-frame point cloud data with semantic information.

\section{Experiment}
\subsection{Semantic Segmentation}
The dataset was utilized to conduct experiments on 2D and 3D semantic segmentation. A total of 13,006 images and 13,006 point cloud frames were collected from 8 short sequences and 1 long sequence for training and testing. The data of scenes 1, 2, 4, 6, 8, 9 were used for model training, and the data of scenes 3, 5, 7 were used for testing. The metrics IoU, mIoU, and mAcc were adopted to evaluate the performance of different semantic segmentation algorithms on this dataset.

\begin{table*}[!t]
\centering
\caption{The quantitative comparison results of five 2D semantic segmentation algorithms.}
\label{table_5}
\begin{tabular}{|c|c|c|c|c|c|c|c|}
\hline
Method       & Lunar regolith & Rock  & Impact crater & Mountain & Sky   & mAcc(\%) & mIoU(\%) \\
\hline
Deeplabv3+   & 98.01          & 75.64 & 79.62         & 83.11    & 99.68 & 92.90    & 87.39    \\
\hline
Mobilenetv3  & 98.72          & 75.16 & 76.96         & 82.98    & 99.57 & 91.99    & 86.68    \\
\hline
Unet         & 98.70          & 73.50 & 73.09         & 79.46    & 99.34 & 91.01    & 84.82    \\
\hline
PointRend    & \textbf{99.07} & 76.17 & \textbf{82.45}& 84.00    & \textbf{99.70} & 93.29    & 88.28    \\
\hline
Segformer    & 99.05          & \textbf{81.69} & 80.37         & \textbf{89.16}    & 99.54 & \textbf{94.22} & \textbf{89.96} \\
\hline
\end{tabular}
\end{table*}

\begin{figure*}[!t]
    \centering
    \subfloat[\label{figure_7_1}]{%
        \includegraphics[width=0.15\textwidth]{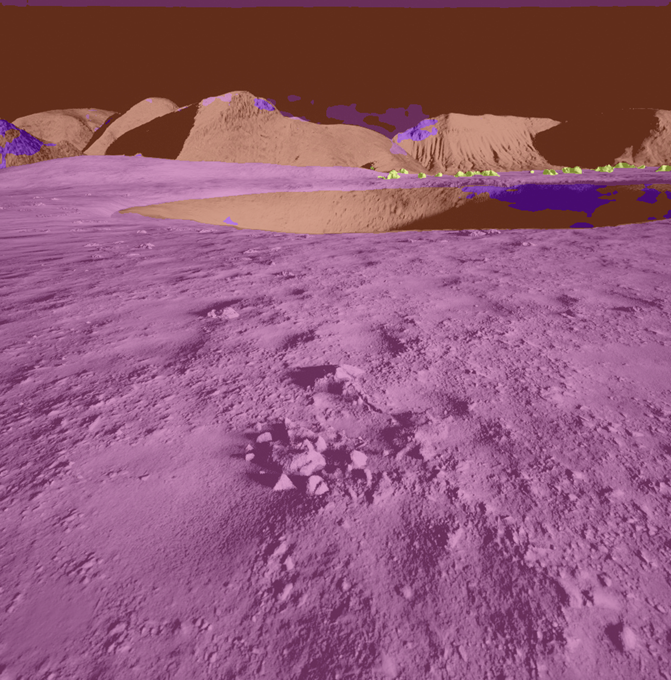}
    }
    \hspace{-1mm}
    \subfloat[\label{figure_7_2}]{%
        \includegraphics[width=0.15\textwidth]{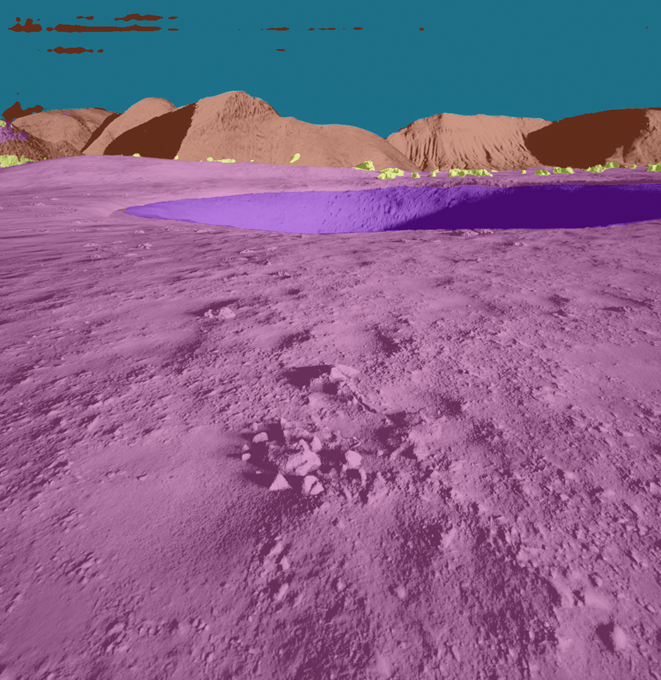}
    }
    \hspace{-1mm}
    \subfloat[\label{figure_7_3}]{%
        \includegraphics[width=0.15\textwidth]{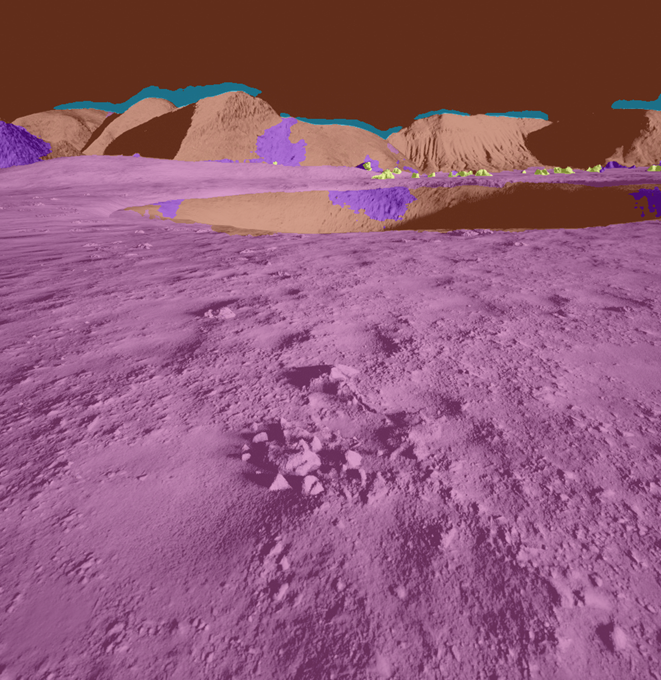}
    }
    \\
    \subfloat[\label{figure_7_4}]{%
        \includegraphics[width=0.15\textwidth]{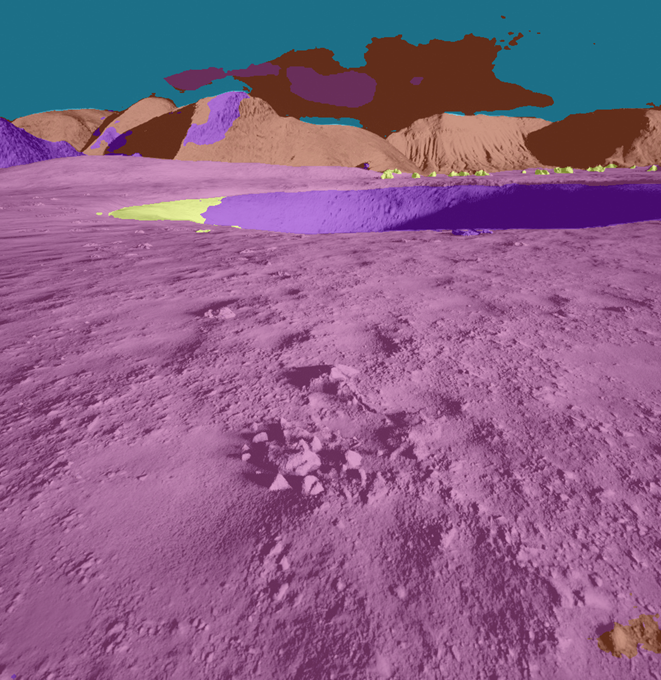}
    }
    \hspace{-1mm}
    \subfloat[\label{figure_7_5}]{%
        \includegraphics[width=0.15\textwidth]{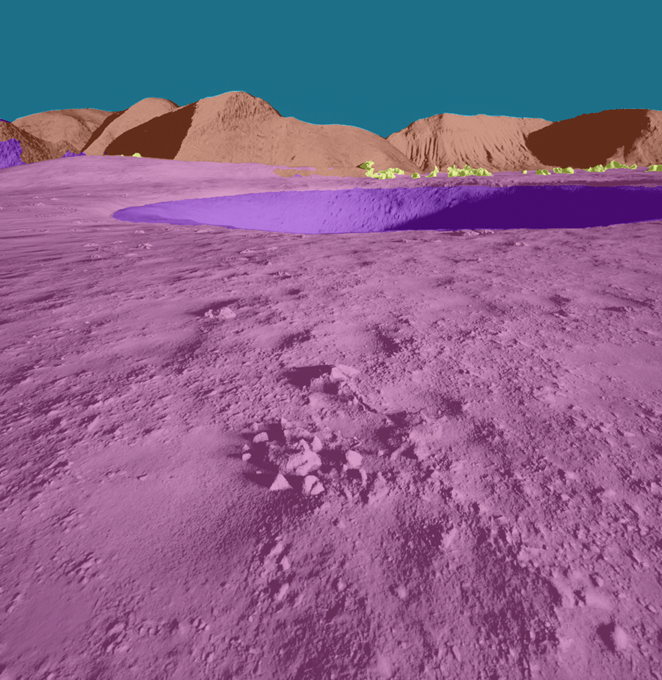}
    }
    \hspace{-1mm}
    \subfloat[\label{figure_7_6}]{%
        \includegraphics[width=0.15\textwidth]{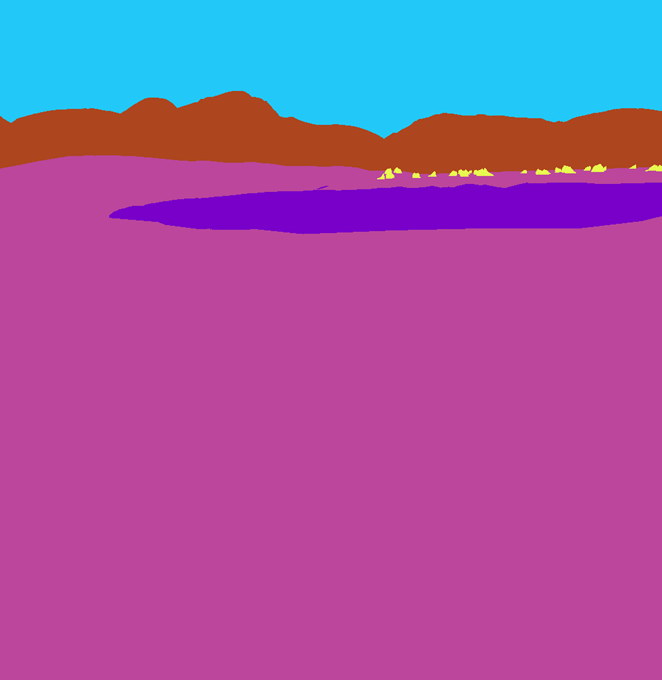}
    }
    \caption{The segmentation visualization of the same image under different algorithms compared with its ground truth. (a) Deeplabv3+ (b) Mobilenetv3 (c) Unet (d) PointRend (e) Segformer (f) Ground truth}
    \label{figure_7}
\end{figure*}

\begin{table*}[h]
\centering
\caption{The quantitative comparison results of five 3D semantic segmentation algorithms.}
\label{table_6}
\begin{tabular}{|c|c|c|c|c|c|}
\hline
Method            & Lunar regolith & Rock          & Impact crater & mAcc(\%) & mIoU(\%) \\
\hline
PointNet          & 96.53          & 76.57         & 11.73         & 66.92    & 61.61    \\
\hline
PointNet++        & 96.69          & 80.80         & 15.03         & 66.80    & 64.17    \\
\hline
KPConv            & 96.09          & 89.95         & 32.04         & 90.94    & 72.69    \\
\hline
RandLA-Net        & 99.56          & 97.65         & 88.51         & \textbf{98.73}    & 95.24    \\
\hline
Point Transformer & \textbf{99.61} & \textbf{98.16} & \textbf{88.67} & 98.09    & \textbf{95.48} \\
\hline
\end{tabular}
\end{table*}

\begin{figure*}[!t]
    \centering
    \subfloat[\label{figure_8_1}]{%
        \includegraphics[width=0.20\textwidth]{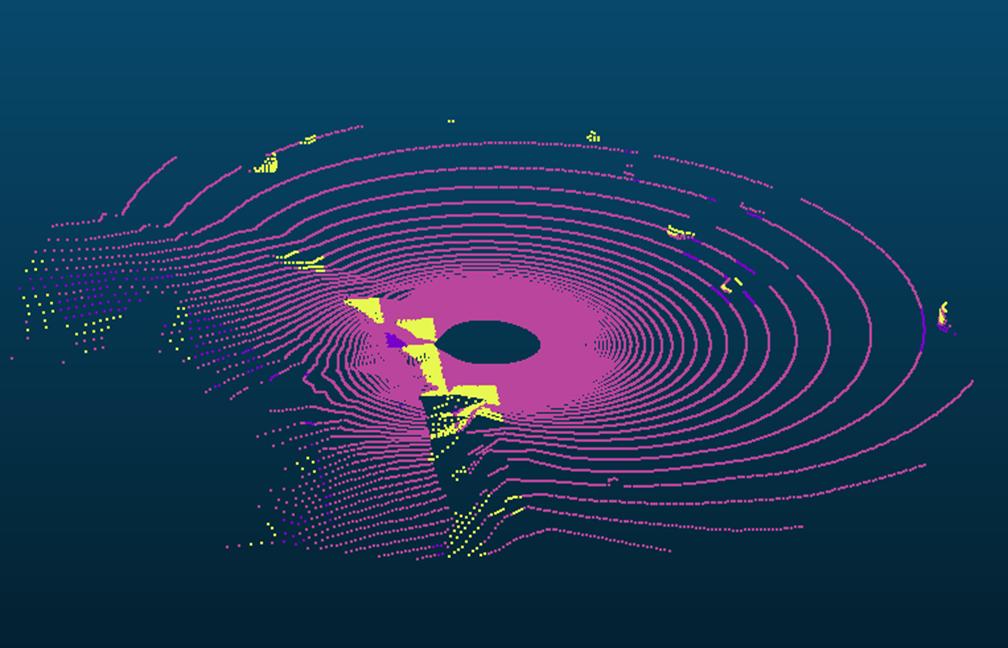}
    }
    \hspace{-1mm}
    \subfloat[\label{figure_8_2}]{%
        \includegraphics[width=0.20\textwidth]{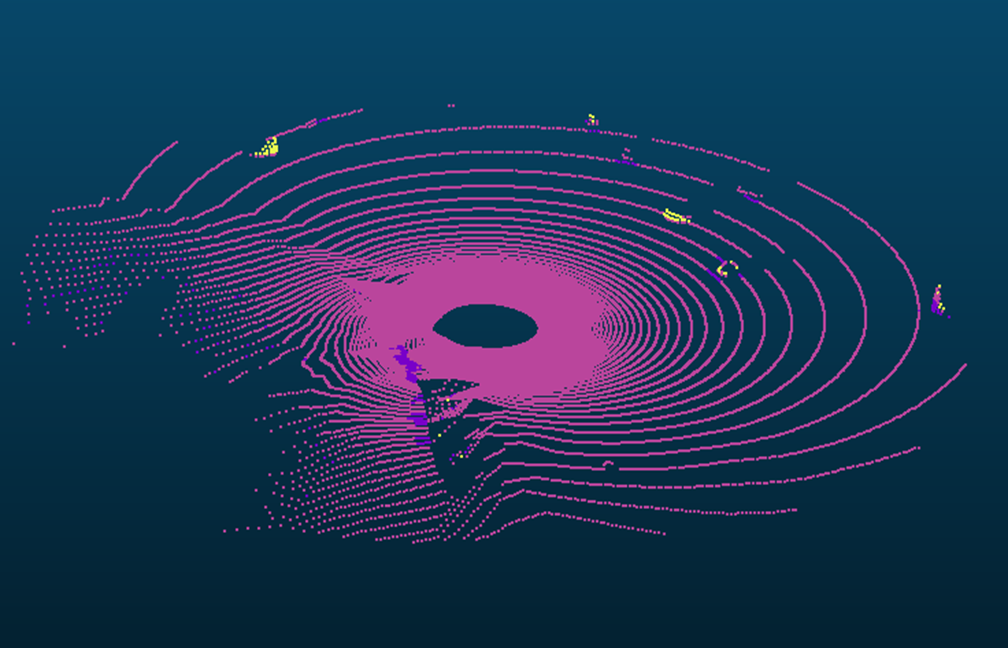}
    }
    \hspace{-1mm}
    \subfloat[\label{figure_8_3}]{%
        \includegraphics[width=0.20\textwidth]{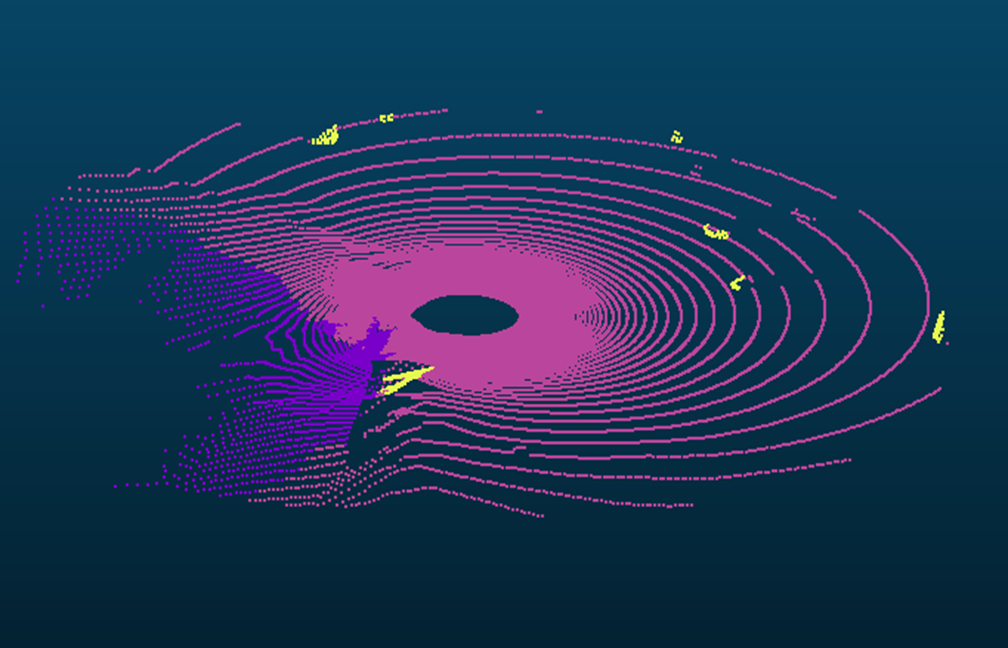}
    }
    \\
    \hspace{-1mm}
    \subfloat[\label{figure_8_4}]{%
        \includegraphics[width=0.20\textwidth]{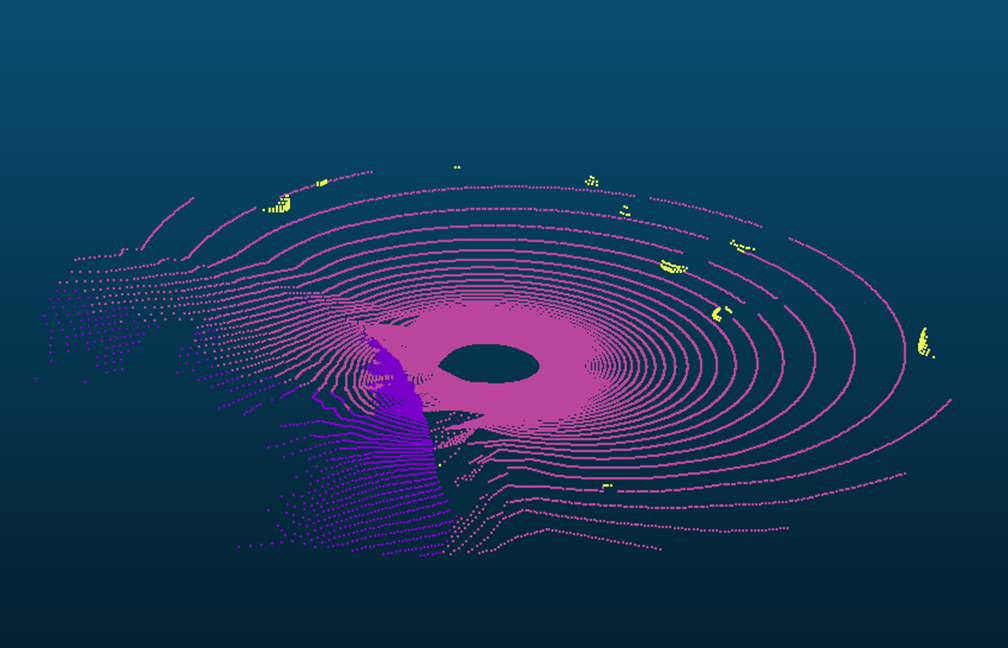}
    }
    \hspace{-1mm}
    \subfloat[\label{figure_8_5}]{%
        \includegraphics[width=0.20\textwidth]{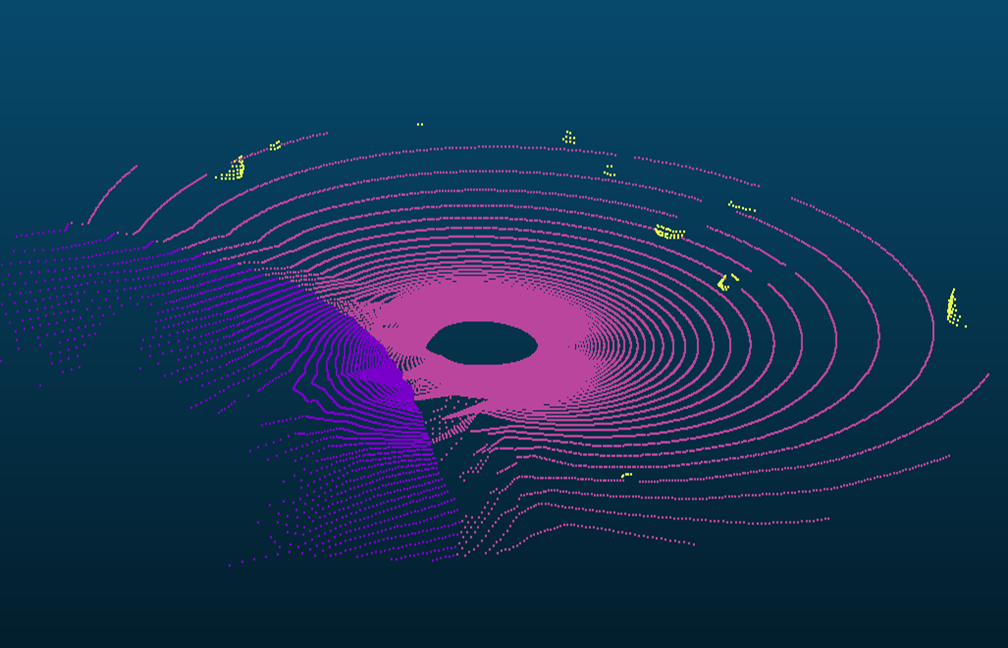}
    }
    \hspace{-1mm}
    \subfloat[\label{figure_8_6}]{%
        \includegraphics[width=0.20\textwidth]{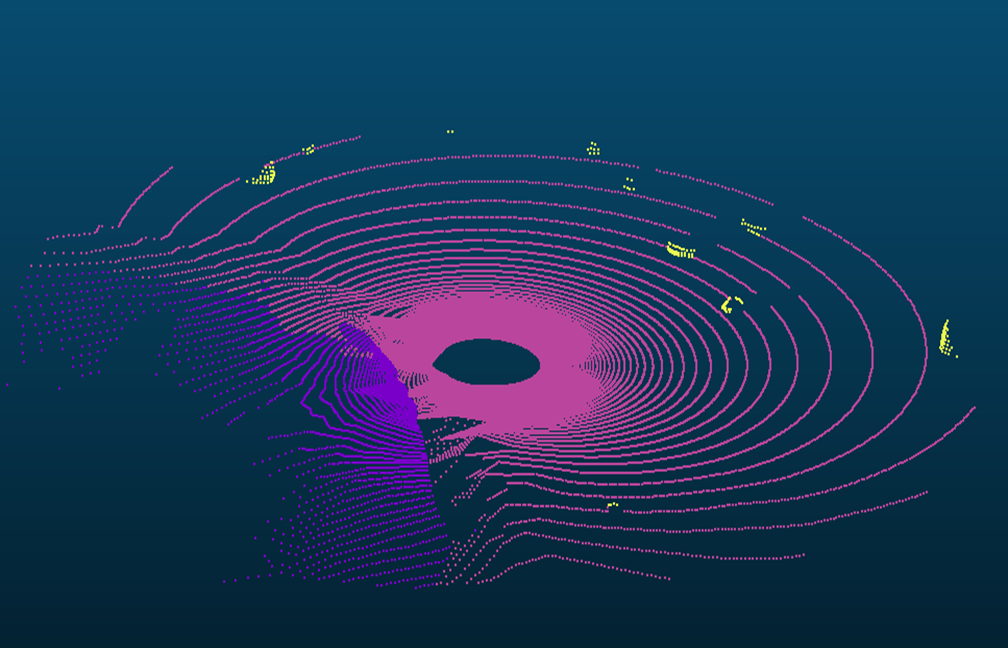}
    }
    \caption{The segmentation visualization of the same point cloud frame under different algorithms compared with its ground truth. (a) PointNet (b) PointNet++ (c) KPConv (d) RandLA-Net (e) Point Transformer (f) Ground truth}
    \label{figure_8}
\end{figure*}

In the 2D semantic segmentation experiment, five sate-of-the-art algorithms are tested: Deeplabv3+~\cite{chen2018encoder}, Mobilenetv3~\cite{howard2019searching}, Unet~\cite{ronneberger2015u}, PointRend~\cite{kirillov2020pointrend}, and Segformer~\cite{xie2021segformer}. The classification categories are lunar regolith, rock, impact crater, mountain, and sky. Table~\ref{table_5} illustrates the quantitative comparison results of five 2D semantic segmentation algorithms. The results presented in Table~\ref{table_5} demonstrate that the PointRend has superior performance in the lunar regolith and impact crater categories, achieving IoU of 99.07\% and 82.45\%. The Segformer outperforms other algorithms in the rock and mountain categories, with an improvement of at least 5\% IoU particularly for the rock and mountain categories. Overall, Segformer has the best segmentation effect on this dataset, as evidenced by its top-ranking mAcc and mIoU of 94.22\% and 89.96\%. Fig.~\ref{figure_7} shows the segmentation visualization of the same image under different algorithms compared with its ground truth. The prediction results of Segformer are basically correct. Mobilenetv3, PointRend and Segformer perform well in identifying impact craters and rock. Both Deeplabv3+ and Unet have errors in the category of impact craters, which may be due to the presence of shadow areas within the craters.

In the 3D semantic segmentation experiment, five sate-of-the-art algorithms are tested: PointNet~\cite{qi2017pointnet}, PointNet++~\cite{qi2017pointnet++}, KPConv~\cite{thomas2019kpconv}, RandLA-Net~\cite{hu2020randla}, and Point Transformer~\cite{zhao2021point}. The classification categories are lunar regolith, rock, and impact crater. Table~\ref{table_6} illustrates the quantitative comparison results of five 3D semantic segmentation algorithms. Fig.~\ref{figure_8} shows the segmentation visualization of the same point cloud frame under different algorithms compared with its ground truth.

It can be seen from Table~\ref{table_6} that Point Transformer consistently achieves optimal performance across all categories, ranking first in mIoU with a tiny advantage of 0.24\%. However, it slightly lags behind RandLA-Net in terms of mAcc by 0.64\%.

By combining Table~\ref{table_5} and Table~\ref{table_6}, it is evident that the 2D semantic segmentation algorithms exhibit excellent segmentation effects on lunar regolith and sky, with IoU consistently exceeding 98\%, while the 3D semantic segmentation algorithms show better results on the categories of close-range rocks and impact crater. Specifically, compared with the best result of the 2D semantic segmentation algorithm, Point Transformer has a higher IoU of 16.47\% for the rock category and a higher IoU of 6.22\% for the impact crater category. It reflects the differences and complementarity of image sequences and point cloud sequences of this dataset in the semantic segmentation tasks. The integration of camera and LiDAR data facilitates more precise segmentation of diverse lunar landforms by rovers, thereby providing richer prior knowledge for path planning and obstacle avoidance.

To test the generalization property of segmentation algorithms, scene 1-3 are selected as training set, and scene 4-9 are used as testing set. The training set and testing set have different data distribution owing to scene design. This experiment will test whether the segmentation algorithm is able to recognize ‘new’ appearance of obstacle which have not seen in the training set.

\begin{table}[h]
\centering
\caption{The quantitative comparison results of Segformer in 6 scenes.}
\label{table_7}
\setlength{\tabcolsep}{5pt} 
\begin{tabular}{|c|c|c|c|c|c|c|c|}
\hline
Scene & Soil  & Rock  & \makecell{Impact \\ crater} & Mountain & Sky   & \makecell{mAcc \\ (\%)} & \makecell{mIoU \\ (\%)} \\
\hline
4     & 98.75 & 87.07 & 66.81         & 96.93    & 99.61 & 92.19    & 89.83    \\
\hline
5     & 98.22 & 70.68 & 60.48         & 74.95    & 99.53 & 89.19    & 80.77    \\
\hline
6     & 98.75 & 85.18 & 68.05         & 93.72    & 99.28 & 92.90    & 89.00    \\
\hline
7     & 98.43 & 86.73 & 66.36         & 93.75    & 99.25 & 92.59    & 88.90    \\
\hline
8     & 98.60 & 83.06 & 73.86         & 86.18    & 98.52 & 92.36    & 88.05    \\
\hline
9     & 96.20 & 76.20 & 19.19         & 76.23    & 96.43 & 78.47    & 72.85    \\
\hline
\end{tabular}
\end{table}

\begin{table}[h]
\centering
\caption{The quantitative comparison results of Point Transformer in 6 scenes.}
\label{table_8}
\setlength{\tabcolsep}{8pt} 
\begin{tabular}{|c|c|c|c|c|c|c|}
\hline
Scene & \makecell{Lunar \\ regolith} & Rock  & \makecell{Impact \\ crater} & \makecell{mAcc \\ (\%)} & \makecell{mIoU \\ (\%)} \\
\hline
4     & 99.83          & 98.87 & 97.74         & 99.31    & 98.81    \\
\hline
5     & 99.49          & 97.91 & 84.04         & 98.12    & 93.81    \\
\hline
6     & 99.31          & 98.41 & 73.27         & 91.76    & 90.33    \\
\hline
7     & 99.63          & 99.05 & 90.41         & 96.94    & 96.36    \\
\hline
8     & 99.54          & 98.38 & 83.83         & 95.07    & 93.92    \\
\hline
9     & 98.24          & 95.94 & 65.06         & 93.98    & 86.41    \\
\hline
\end{tabular}
\end{table}

Table~\ref{table_7} and Table~\ref{table_8} illustrate the quantitative comparison results of Segformer and Point Transformer tested in different scenes. As can be seen that Segformer achieves an average mIoU of 84.90\% across six scenes, while Point Transformer attains an average mIoU of 93.27\%. Despite being trained solely on simple scene data, both algorithms achieve a certain level of generalization in complex environments. The 3D semantic segmentation algorithm outperforms its 2D counterpart in terms of generalization ability, showcasing an average mIoU improvement of 8.37\%.

\begin{figure}[h]
\centering
\includegraphics[width=\columnwidth]{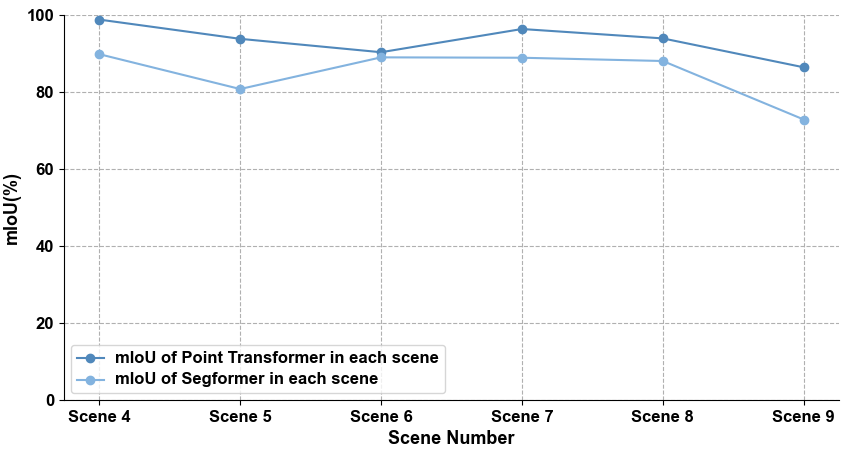}
\caption{The Line chart of mIoU of the two 2D/3D algorithms in 6 scenes.}
\label{figure_9}
\end{figure}

Fig.~\ref{figure_9} demonstrates the variations of mIoU for 2D and 3D segmentation algorithms in the six scenes. The segmentation effect is the best in Scene 4, but it is the worst in Scene 9. In terms of topographic relief, scenes 4, 5, and 6 roughly show a decreasing trend in mIoU as the slope steepness rises, as do scenes 7, 8, and 9. In scenes 5 and 6, compared to scene 4, the mIoU of the Segformer decreases with increasing topographic relief, with scene 5 having the lowest accuracy. This may be due to the presence of extensive shadow areas in scene 5. In scenes 7, 8, and 9, as topographic relief increases, the mIoU of the segformer also declines. The mIoU of the Point Transformer shows a decreasing trend in scenes 4, 5, 6, and 7, 8, and 9, respectively. Regarding the density of objects, Segformer's average mIoU is 85.30\% in scenes 4-6 and drops to 80.87\% in scenes 7-9; Point Transformer’s average mIoU is 94.32\% in scenes 4-6 and drops to 92.23\% in scenes 7-9. It is interesting to see that with the increase of lunar surface objects richness, the segmentation performance decreases. The findings suggest that although the segmentation algorithms do show a certain level of generalization, their performance in semantic segmentation of the lunar surface is limited due to significantly influenced by topographic relief and the density of objects. Therefore, further research can be carried on to enhance the generalization property of segmentation for lunar related applications.

\subsection{SLAM}
Visual SLAM and LiDAR SLAM experiments were conducted using this dataset. Since the dataset provides the ground truth of the sensor's pose in the world frame, and the SLAM estimates the pose relative to the first frame in the local odometry frame based on the sensor. In order to accurately evaluate the performance of SLAM algorithm , it is necessary to use the extrinsic of the sensor relative to the rover and the real-time ground truth pose of the rover in the world frame provided by the dataset to convert the estimated relative pose to the world frame. The conversion relationship is as follows:

\begin{equation}
\label{equation_1}
T_{b_{0}}^{b_{i}} = T_{b}^{o} T_{o_{0}}^{o_{i}} T_{o}^{b}.
\end{equation}

In the equation \eqref{equation_1}, $T_{o_{0}}^{o_{i}}$ indicates the pose of odometry at the timestamp $i$ relative to the starting point of the motion at the timestamp 0, which is the value estimated by the SLAM. $T_{b}^{o}$ is the transformation matrix from the odomoetry frame to the rover’s native frame obtained based on the extrinsic of the sensor. $T_{o}^{b}$ is the inverse matrix of $T_{b}^{o}$, which denotes the transformation matrix from the rover’s native frame to the odometry frame. According to equation \eqref{equation_1}, $T_{o_{0}}^{o_{i}}$ can be converted to the pose $T_{b_{0}}^{b_{i}}$ of the rover's native frame at the timestamp $i$ relative to the rover's native frame at the timestamp 0. For the convenience of representation, the following definitions are made:

\begin{equation}
\label{equation_2}
P_i := T_{b_{0}}^{b_{i}}.
\end{equation}

\begin{equation}
\label{equation_3}
Q_i := T_w^{b_{i}}.
\end{equation}

The estimated trajectory of SLAM is $P_1$, $P_2$, ..., $P_n \in SE(3)$ and the ground truth trajectory is $Q_1$, $Q_2$, ..., $Q_n \in SE(3)$. There are two main metrics for evaluation: relative pose error (RPE) and absolute trajectory error (ATE)~\cite{sturm2012benchmark}. The RPE at timestamp $i$ is defined as follows:

\begin{equation}
\label{equation_4}
E_i := \left( Q_i^{-1} Q_{i+\Delta} \right)^{-1} \left( P_i^{-1} P_{i+\Delta} \right).
\end{equation}

The odometry of matching between consecutive frames is suitable for calculating the root mean square error (RMSE)~\cite{sturm2012benchmark} in the case of $\Delta = 1$. That is:

\begin{equation}
\label{equation_5}
RMSE(E_{1:n}, 1) := \left( \frac{1}{n-1} \sum_{i=1}^{n-1} \| \text{trans}(E_i) \|^2 \right)^{\frac{1}{2}}.
\end{equation}

where $\text{trans}(E_i)$ indicates the translational components of the relative pose error $E_i$.

The ATE at timestamp $i$ is defined as follows:

\begin{equation}
\label{equation_6}
F_i := Q_i^{-1} S P_i.
\end{equation}

where $S$ refers to the transformation between the estimated trajectory and the ground truth trajectory for alignment. In the past, the method of Horn~\cite{sturm2012benchmark,horn1987closed} or Umeyama~\cite{umeyama1991least} was usually used, but it is not able to evaluate the accumulative error. Therefore, the calculated transformation $S$ is used to convert $P_i$ into the same frame as $Q_i$, in orde to more accurately evaluate the performance of SLAM. The formula for the conversion is as follows: 

\begin{equation}
\label{equation_7}
S = T_{w}^{b_i} = T_{w}^{b_0} T_{b_0}^{b_i}.
\end{equation}

Apparently, $S$ is the rover’s pose under the world frame at timestamp $i$. Similarly, RMSE is used as metric, that is:

\begin{equation}
\label{equation_8}
RMSE(F_{1:n}) := \left( \frac{1}{n} \sum_{i=1}^{n} \| \text{trans}(F_i) \|^2 \right)^{\frac{1}{2}}.
\end{equation}

Considering the influence of trajectory length on the RMSE of ATE, where shorter trajectories exhibit less translational and rotational drift~\cite{giubilato2022challenges}. In order to fairly compare all scenes, the percentage of the RMSE of RPE and ATE relative to the trajectory length is calculated as the evaluation metrics, as follows:

\begin{equation}
\label{equation_9}
RMSE_{RPE} = RMSE(E_{1:n}, 1) \, m.
\end{equation}

\begin{equation}
\label{equation_10}
RMSE_{ATE} = \frac{100 \times RMSE(F_{1:n})}{length_{gt}} \%.
\end{equation}

ORB-SLAM3~\cite{campos2021orb} (stereo module and stereo inertial module), VINS-Mono~\cite{qin2018vins}, and VINS-Fusion~\cite{qin2019generala,qin2019generalb} are selected for visual SLAM, while A-LOAM~\cite{zhang2014loam}, LeGO-LOAM~\cite{shan2018lego} and FAST-LIO~\cite{xu2022fast} are selected for LiDAR SLAM, where FAST-LIO is a multi-source sensor algorithm of the tightly coupled IMU. 

\begin{table*}[!t]
\centering
\caption{The accuracy of visual SLAM and LiDAR SLAM algorithms in 9 scenes.}
\label{table_9}
\setlength{\tabcolsep}{2.5pt} 
\begin{tabular}{|c|c|c|c|c|c|c|c|c|c|c|c|c|c|c|c|c|}
\hline
\multirow{2}{*}{\raisebox{-2.0\height}{Scene}} & \multicolumn{8}{c|}{Vision SLAM} & \multicolumn{6}{c|}{LiDAR SLAM} \\
\cline{2-15}
 & \multicolumn{2}{c|}{ORB-SLAM3(s)} & \multicolumn{2}{c|}{ORB-SLAM3(si)} & \multicolumn{2}{c|}{VINS-Mono} & \multicolumn{2}{c|}{VINS-Fusion} & \multicolumn{2}{c|}{A-LOAM} & \multicolumn{2}{c|}{LeGO-LOAM} & \multicolumn{2}{c|}{FAST-LIO} \\
\cline{2-15}
 & \scriptsize RMSE$_{\scriptsize \text{RPE}}$ & \scriptsize RMSE$_{\scriptsize \text{ATE}}$ & \scriptsize RMSE$_{\scriptsize \text{RPE}}$ & \scriptsize RMSE$_{\scriptsize \text{ATE}}$ & \scriptsize RMSE$_{\scriptsize \text{RPE}}$ & \scriptsize RMSE$_{\scriptsize \text{ATE}}$ & \scriptsize RMSE$_{\scriptsize \text{RPE}}$ & \scriptsize RMSE$_{\scriptsize \text{ATE}}$ & \scriptsize RMSE$_{\scriptsize \text{RPE}}$ & \scriptsize RMSE$_{\scriptsize \text{ATE}}$ & \scriptsize RMSE$_{\scriptsize \text{RPE}}$ & \scriptsize RMSE$_{\scriptsize \text{ATE}}$ & \scriptsize RMSE$_{\scriptsize \text{RPE}}$ & \scriptsize RMSE$_{\scriptsize \text{ATE}}$ \\
\hline
1 & 0.0046 & 0.1933 & 0.0038 & 0.1411 & 0.0527 & 3.8933 & 1.3211 & 2.7246 & 0.3215 & 1.2503 & 4.3903 & 6.8975 & 0.0720 & 4.3340 \\
\hline
2 & 0.0019 & 0.0891 & 0.0044 & 0.0755 & 0.0307 & 1.2133 & 0.6186 & 0.9744 & 0.5900 & 1.5498 & 7.2205 & 4.2094 & 0.0439 & 0.1898 \\
\hline
3 & 0.0020 & 0.1456 & 0.0040 & 0.0619 & 0.0211 & 2.1130 & 0.4216 & 0.6574 & 0.2924 & 0.4387 & 7.1883 & 7.8525 & 0.0487 & 1.7975 \\
\hline
4 & 0.0024 & 0.0554 & 0.0055 & 0.0409 & 0.0269 & 1.8580 & 0.4262 & 1.1383 & 0.7307 & 2.1245 & 2.3536 & 28.7922 & 0.0432 & 0.4615 \\
\hline
5 & 0.0020 & 0.0444 & 0.0070 & 0.0773 & 0.0278 & 1.7708 & 0.1306 & 0.8625 & 0.3340 & 1.7731 & 2.8573 & 13.1838 & 0.0575 & 1.5479 \\
\hline
6 & 0.0043 & 0.4182 & 0.0240 & 0.4632 & 0.0161 & 1.3064 & 0.1726 & 0.5807 & 0.2914 & 1.3599 & 8.8268 & 9.8600 & 0.0531 & 0.5209 \\
\hline
7 & 0.0017 & 0.0745 & 0.0052 & 0.0466 & 0.0363 & 2.4146 & 0.4961 & 0.6831 & 0.3513 & 1.3278 & 4.1259 & 6.1484 & 0.0388 & 0.3577 \\
\hline
8 & 0.0018 & 0.0483 & 0.0070 & 0.0416 & 0.0335 & 2.5308 & 0.6506 & 1.6375 & 0.7202 & 11.6698 & 7.0844 & 16.9650 & 0.0468 & 0.6867 \\
\hline
9 & 0.0019 & 0.1132 & 0.0061 & 0.0968 & 0.0510 & 0.7134 & 0.2398 & 0.8163 & 0.2681 & 2.1814 & 0.7021 & 3.8255 & 0.0449 & 2.3525 \\
\hline
\end{tabular}
\end{table*}

\begin{figure*}[!t]
\centering
\includegraphics[width=0.85\textwidth]{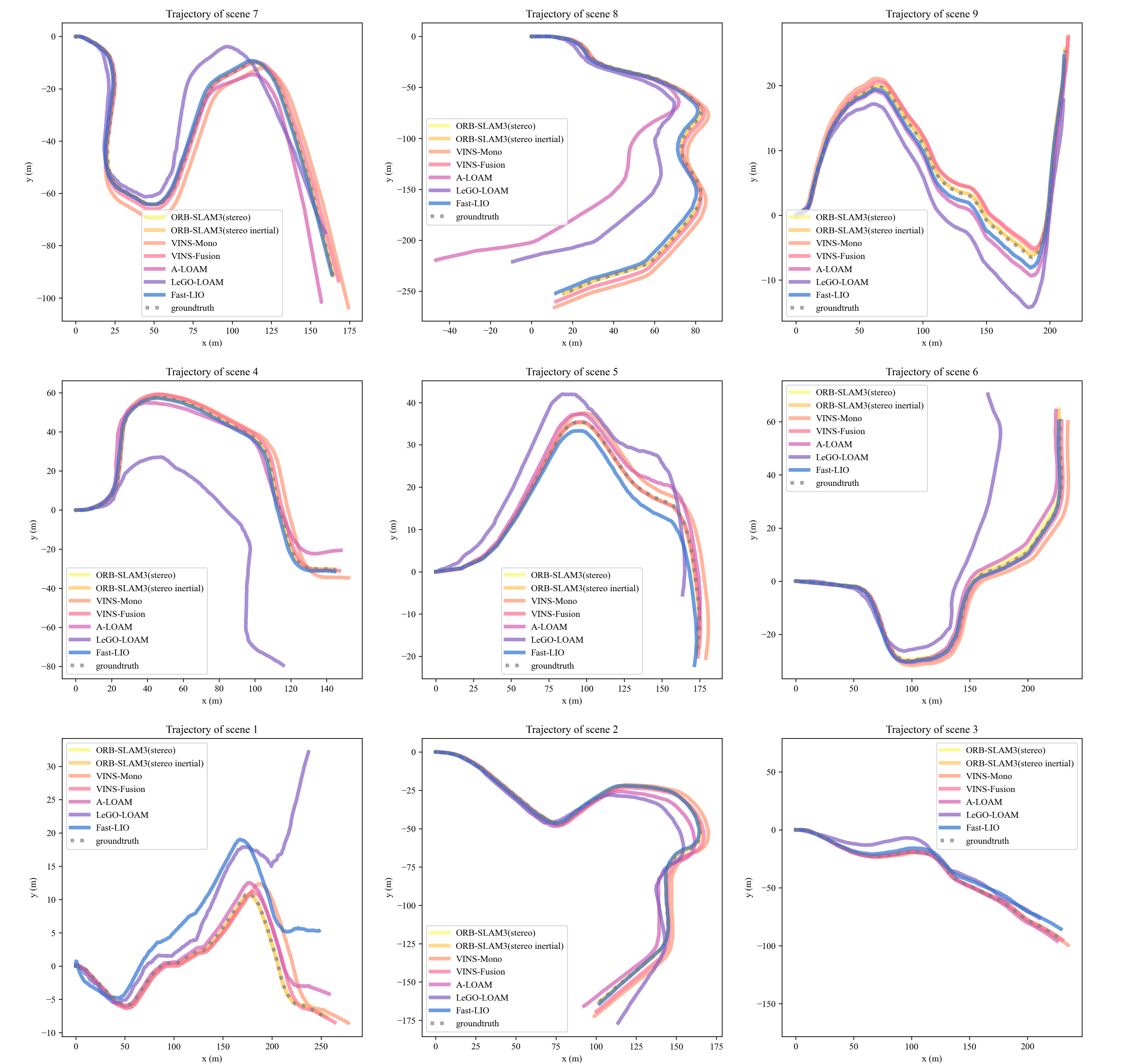}
\caption{Trajectories visualization of visual odometry and LiDAR odometry in 9 scenes.}
\label{figure_10}
\end{figure*}

Table~\ref{table_9} summaries the accuracy of the above algorithms in 9 scenes and Fig.~\ref{figure_10} shows the trajectories visualization of visual odometry and LiDAR odometry in 9 scenes. For the visual SLAM experiment, combining the results of 9 scenes, the average $RMSE_{RPE}$ of ORB-SLAM3 in stereo mode and stereo inertial mode are 0.0025 and 0.0074, and the average $RMSE_{ATE}$ are 0.1313\% and 0.1161\%, respectively. The average $RMSE_{RPE}$ of VINS-Mono and VINS-Fusion are 0.0329 and 0.4975, and the average $RMSE_{ATE}$ are 1.9793\% and 1.1194\%, respectively. It can be seen that the benefits of ORB feature points for scene feature extraction are obvious, as evidenced by the ORB-SLAM3 has a higher accuracy than VINS-Mono and VINS-Fusion based on optical flow features. And the addition of IMU measurements makes ORB\_SLAM3 in stereo inertial mode has a lower $RMSE_{ATE}$ compared to ORB-SLAM3 in stereo mode. For the LiDAR SLAM experiment, it can be seen that FAST-LIO has the best overall performance in 9 scenes due to the combination of IMU. In terms of $RMSE_{ATE}$, LeGO-LOAM with the highest dependence on ground objects performs poorly, with an average $RMSE_{ATE}$ of 10.8594\% in 9 scenes. A-LOAM demonstrates moderate performance, with an average $RMSE_{ATE}$ of 2.6306\%, and FAST-LIO performs better, with $RMSE_{ATE}$ reduced by 9.4985\% and 1.2697\% compared to LeGO-LOAM and ALOAM.

\begin{figure*}[!t]
    \centering
    \subfloat[\label{figure_11_1}]{
        \includegraphics[width=0.38\textwidth]{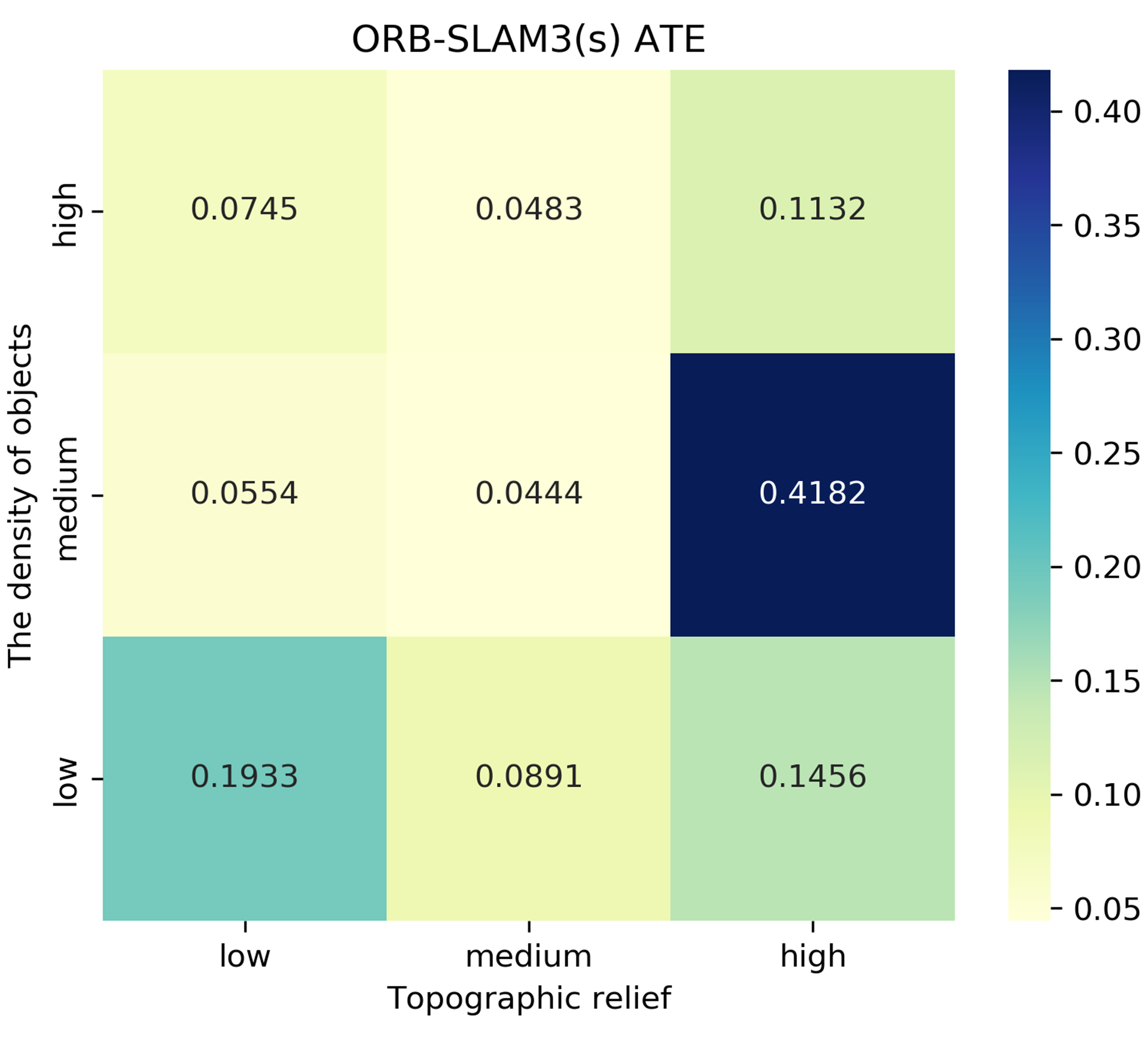}
    }
    \subfloat[\label{figure_11_2}]{
            \includegraphics[width=0.38\textwidth]{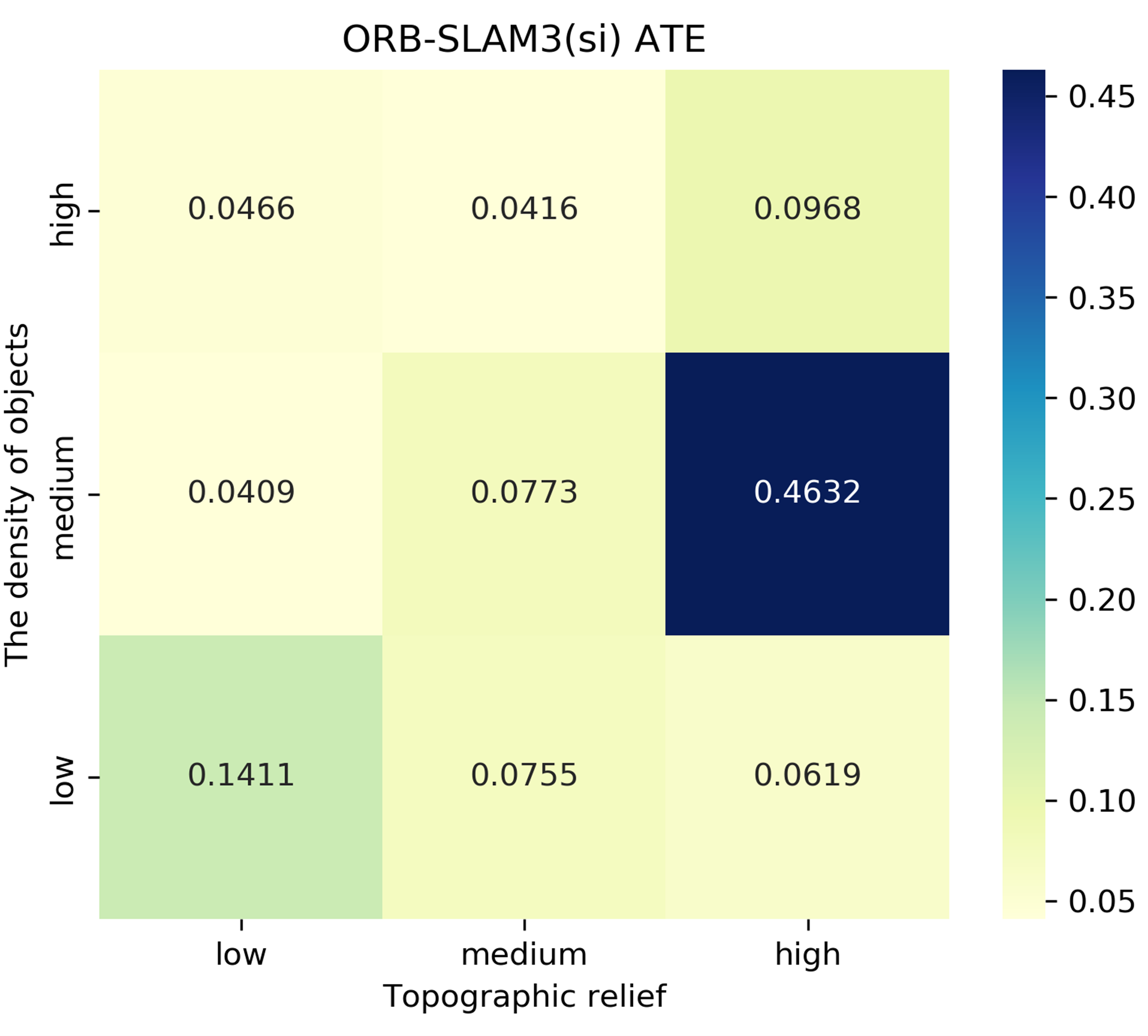}
    }  
    \\
    \subfloat[\label{figure_11_3}]{
        \includegraphics[width=0.38\textwidth]{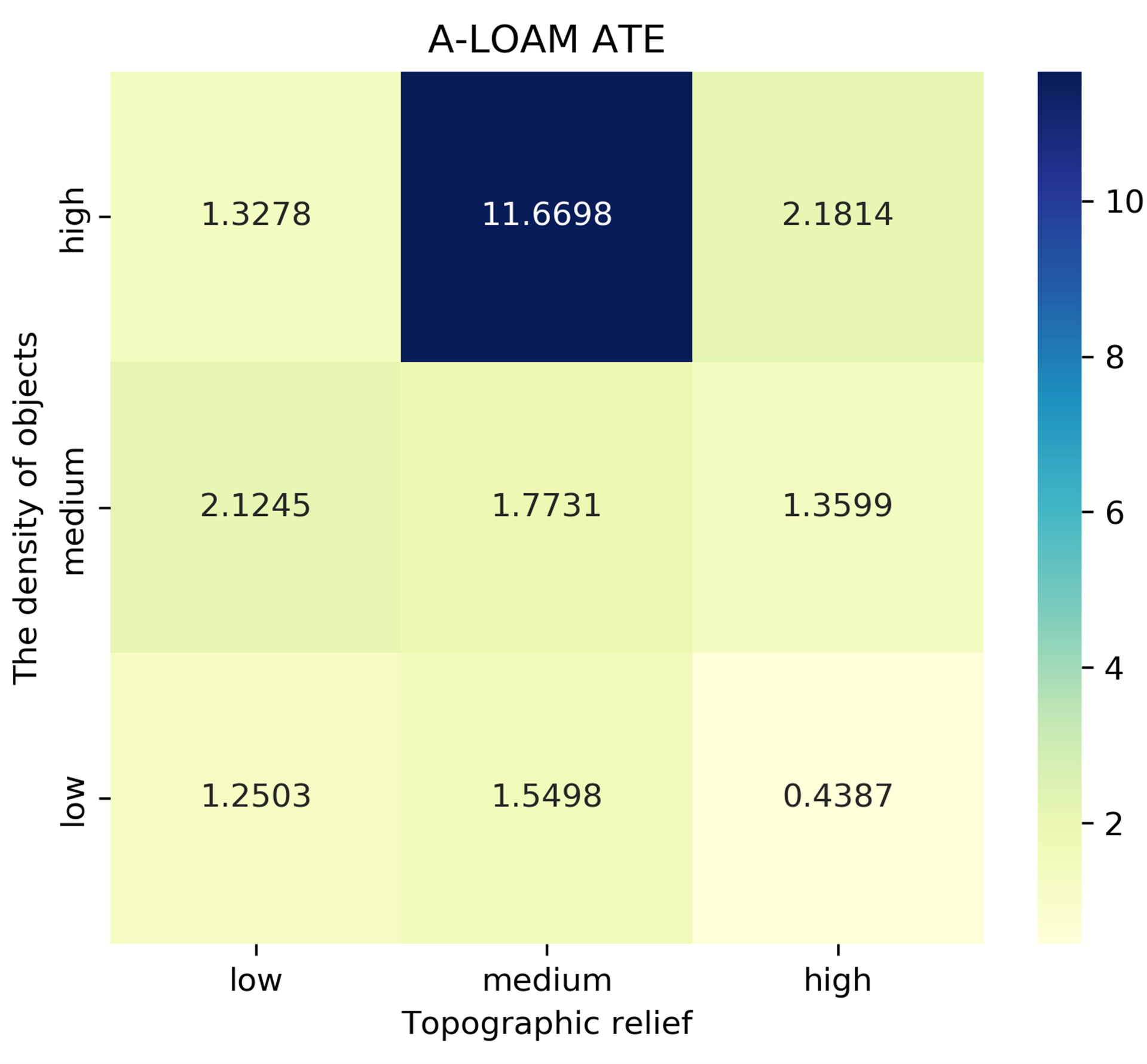}
    }
    \subfloat[\label{figure_11_4}]{
        \includegraphics[width=0.38\textwidth]{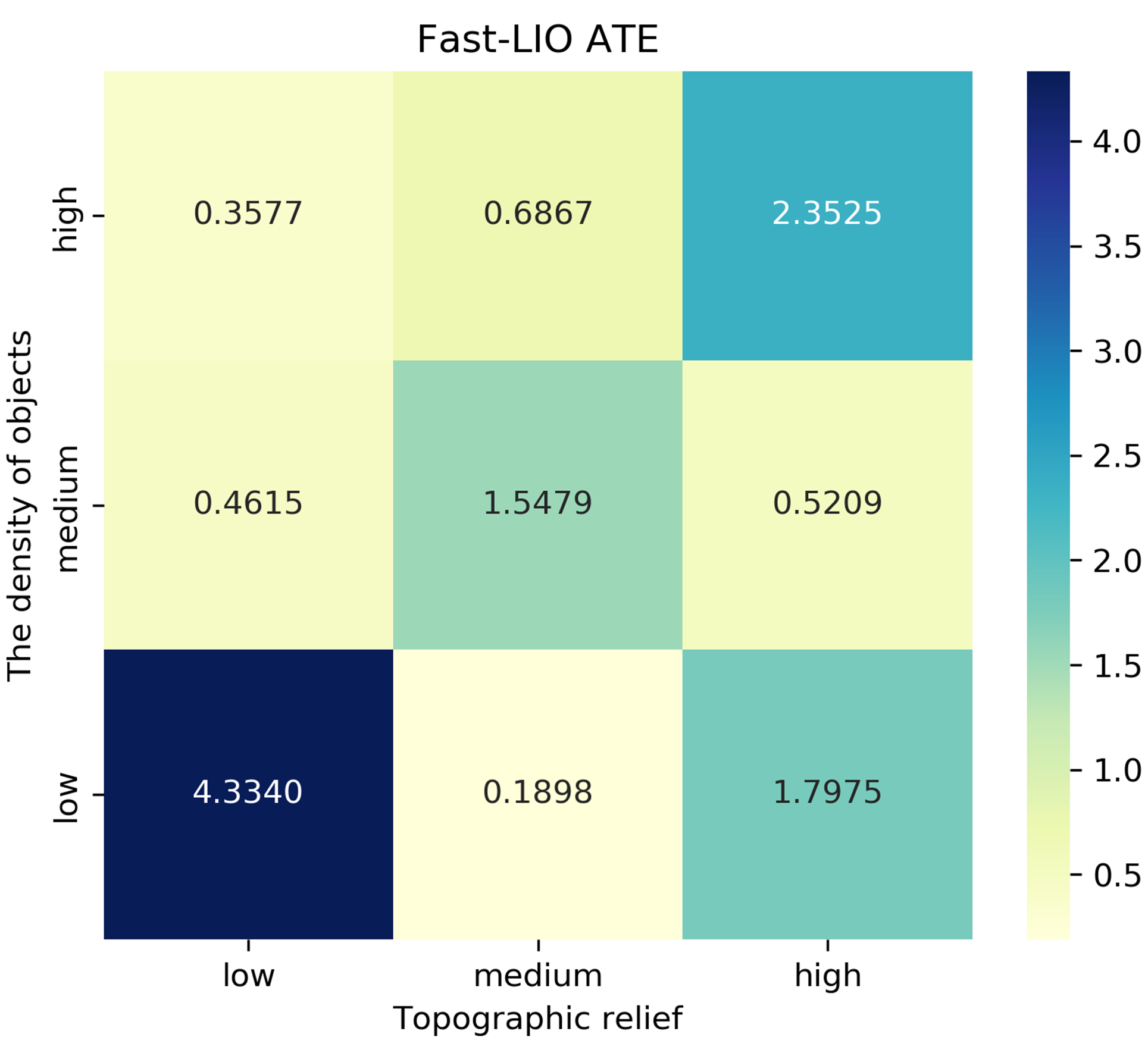}
    }
    \caption{The heatmap of visual odometry and LiDAR odometry. (a)ORB-SLAM3\_stereo (b)ORB-SLAM3\_stereo\_inertial (c)A-LOAM (d)Fast-LIO}
    \label{figure_11}
\end{figure*}

In order to more intuitively analyze the impact of topographic relief and the density of objects on different types of SLAM algorithms, the $RMSE_{ATE}$ of ORB-SLAM3 (stereo) in visual odometry, ORB-SLAM3 (stereo inertial) in visual inertial odometry, LiDAR odometry A-LOAM, and LiDAR inertial odometry FAST-LIO are selected and plotted in a heatmap as shown in Fig.~\ref{figure_11}, with darker colors denoting larger errors.

From Fig.~\ref{figure_11}\subref{figure_11_1}, it can be observed that the visual odometry   generally decreases as the density of objects increases, reaching the minimum value of 0.0444\% in scene 5, and the value of scene 8 is only slightly higher than scene 5 by 0.0041\%; from Fig.~\ref{figure_11}\subref{figure_11_2}, it can be observed that the visual inertial odometry $RMSE_{ATE}$ also decreases with increasing objects density, reaching the minimum value of 0.0416\% in scene 8. The findings suggest that visual SLAM requires higher density of objects and proper topographic relief, as an increased object density facilitates improved extraction of image features, while proper topographic relief enriches image features by reducing occlusion.

From Fig.~\ref{figure_11}\subref{figure_11_3}, it can be seen that the performance of LiDAR odometry in scene 8 is very poor, reaching $RMSE_{ATE}$ of 11.6698\%, and the average $RMSE_{ATE}$ of the 3 scenes with different obejects density decreases as the level increases; from Fig.~\ref{figure_11}\subref{figure_11_4}, it can be seen that the LiDAR inertial odometry has the worst effect in the simplest scene 1, with $RMSE_{ATE}$ of 4.3340\%. Meanwhile, it also does not perform well in the moset complex scene 9, with $RMSE_{ATE}$ of 2.3525\%. The findings suggest that LiDAR SLAM heavily relies on appropriate combination of lunar topographic relief and objects density, otherwise the results will be poor.

Overall, the highest $RMSE_{ATE}$ of visual SLAM is 0.4632\%, while the average $RMSE_{ATE}$ of LiDAR SLAM is 11.9958\%, thereby the LiDAR SLAM is worse than visual SLAM in this dataset. Consequently, the visual SLAM is more suitable for navigation tasks in luanr scenes.

The results of $RMSE_{ATE}$ after adding the IMU measurements to the odometery demonstrate that the accuracy has improved except for scenes 5 and 6 in visual SLAM, with an average $RMSE_{ATE}$ of 0.0307\% reduced; in the LiDAR SLAM, except for scenes 1, 3, and 9, the accuracy of the scenes has improved, with an average $RMSE_{ATE}$ of 2.67\% reduced. It shows that IMU has certain ability to improve navigation accuracy in lunar scenes.

\subsection{3D Reconstruction}
The stereo matching and 3D reconstruction algorithms based on the LuSNAR dataset were tested. In practical detection tasks, it is often necessary to quickly generate depth information for unknown environments to support decision-making. This requires the inspector to be able to perform zero sample disparity estimation without data support, and to have the ability to quickly understand and adapt to unknown scenes. Therefore, this paper tested the zero sample depth estimation ability of various stereo matching models based on the proposed Lusnar dataset.

In the stereo matching experiment, 10 pairs of stereo images that can reflect the representativeness of the scene were selected from 9 sequences, BM and SGBM~\cite{ueshiba2006efficient}, PSMNet~\cite{chang2018pyramid}, RAFTStereo~\cite{lipson2021raft} and CREStereo~\cite{li2022practical} were used for disparity estimation. Among them, BM and SGBM were configured with a maximum disparity of 256 and a window size of 7, while PSMNet, RAFT-Stereo and CREStereo were tested with pre-trained models.

\begin{table}[h]
\centering
\caption{Experimental results of stereo matching on LuSNAR dataset}
\label{table_10}
\setlength{\tabcolsep}{2pt} 
\begin{tabular}{|c|c|c|c|c|c|}
\hline
\diagbox{Scene}{Method} & BM & SGBM & PSMNet & RAFTStereo & CREStereo \\ \hline
1 & 38.30 & 76.78 & 62.93 & 23.63 & \textbf{20.04} \\ \hline
2 & 33.50 & 71.20 & 62.86 & 23.66 & \textbf{18.09} \\ \hline
3 & 31.98 & 70.75 & 59.67 & 26.57 & \textbf{16.92} \\ \hline
4 & 33.67 & 69.90 & 64.59 & 28.95 & \textbf{17.44} \\ \hline
5 & 35.91 & 71.41 & 64.28 & 23.84 & \textbf{5.82} \\ \hline
6 & 74.07 & 74.06 & 63.16 & 21.77 & \textbf{20.95} \\ \hline
7 & 30.02 & 71.02 & 62.72 & 26.16 & \textbf{27.01} \\ \hline
8 & 32.73 & 71.15 & 62.88 & 22.79 & \textbf{12.20} \\ \hline
9 & \textbf{64.62} & 71.97 & 98.16 & 83.12 & 69.83 \\ \hline
\makecell{Average \\ Error Rate} & 41.64 & 72.02 & 66.80 & 31.16 & \textbf{23.14} \\ \hline
\end{tabular}
\end{table}

\begin{figure}[!t]
\centering
\includegraphics[width=\columnwidth]{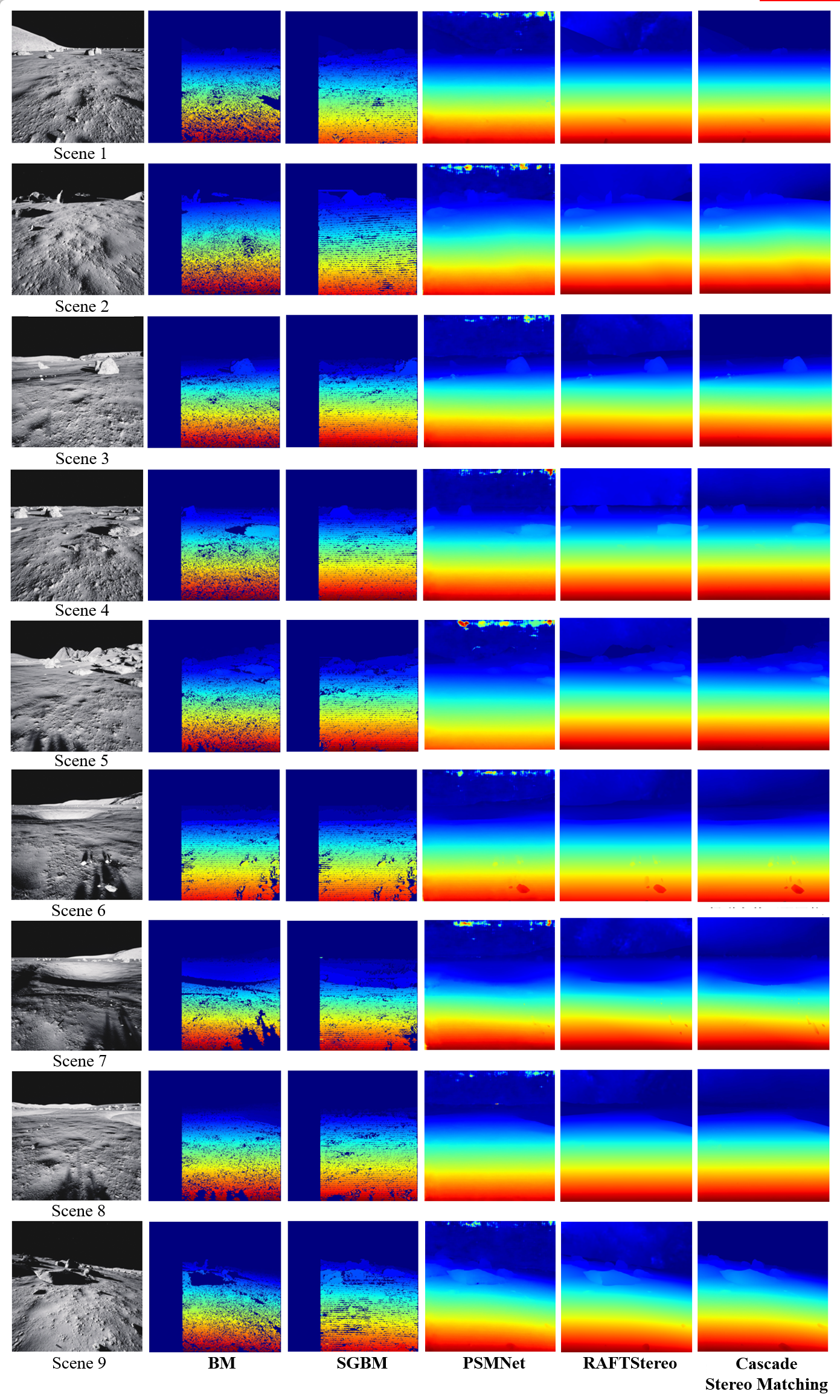}
\caption{Visualization of stereo matching results from LuSNAR dataset.}
\label{figure_12}
\end{figure}

The evaluation metric for stereo matching accuracy is the pixel error. When the error between the predicted disparity and the ground truth disparity exceeds 1 pixel, it is considered a matching failure. Lower pixel error indicates better stereo matching performance. Table~\ref{table_10} illustrates the quantitative comparison results of five stereo matching algorithms. The results presented in it demonstrate that CREStereo has superior overall performance, with an average error rate of 23.14\% across 9 scenes. RAFTStereo slightly outperforms CREStereo by 0.85\% in terms of error rate in scene 7, while BM works best in scene 9 characterized by the most complex terrain, exhibiting an error rate that is 5\% lower than that of CREStereo. The difficulty of stereo matching in scene 9 is that there are a large number of shadow areas. BM algorithm is relatively simple, it estimates disparity by finding the best matching block. This simplicity may mean fewer assumptions and smaller error accumulation.

Fig.~\ref{figure_12} shows the visualization of disparity generated by five algorithms in scenarios 1-9. The color change from red to blue represents the distance from near to far. It can be seen from the disparity map that the disparity map predicted by BM and sgbm as traditional methods is sparse and noisy, while the disparity results obtained from psmnet, raft stereo and CREStereo are more smooth. Specifically, the CREStereo can save more scene details. For example, in Scene 3, the debris near the edge of the impact crater and the crater; Scene 5 rocks on the far slope; The result of CREStereo is more refined than other algorithms. From the disparity map visualization, it can be seen that the reason for the mismatch is largely due to the difficulty in correctly evaluating the disparity between the sky and distant mountains. The true values of the depths of the sky and mountains range from tens of meters to hundreds of meters. In order to show the depth estimation ability of stereo matching network for the whole scene range, the experimental results also include the errors caused by the sky category.

The analysis of CREStereo results based on topographic relief and feature richness shows that the average mismatch rate of scenes 2, 5 and 8 with moderate topographic relief is 9.46\% lower than that of scenes 1, 4 and 7 with low topographic relief, and 23.87\% lower than that of scenes 3, 6 and 9 with high topographic relief. Similarly, the average mismatching rate of scenes 4, 5 and 6 with medium ground feature richness is 3.62\% and 21.61\% lower than that of scenes 1, 2 and 3 with low ground feature richness and scenes 7, 8 and 9 with high ground feature richness, respectively. It can be seen that high topographic relief and objects density have a significant impact on the accuracy of stereo matching. It is possible that mountains or large rocks block the light, resulting in shaded areas. These areas lack texture features, thereby hindering the accurate estimation of disparity by stereo matching algorithms. The shadow area in scene 9, for instance, has the difficulties in estimating disparity based on visual texture due to its potential confusion with the black sky. Consequently, the mismatch rate of scene 9 surpasses that of other scenes. In general, medium topographic relief and objects density can provide appropriate texture information for stereo matching, minimizing the presence of excessive shadows and facilitating the acquisition of more accurate disparity.

\begin{table}[h]
\centering
\caption{Experimental results of 3D reconstruction of LuSNAR dataset}
\label{table_11}
\begin{tabular}{|c|c|c|c|c|}
\hline
\diagbox{Scene}{Range} & 5m & 10m & 20m & 50m \\ \hline
1 & 0.0081 & 0.0106 & 0.2175 & 0.6976 \\ \hline
2 & 0.0052 & 0.0086 & 0.3196 & 1.0175 \\ \hline
3 & 0.0058 & 0.0097 & 0.3119 & 0.9536 \\ \hline
4 & 0.0050 & 0.0087 & 0.4754 & 1.1130 \\ \hline
5 & 0.0068 & 0.0105 & 0.0264 & 0.4143 \\ \hline
6 & 0.0047 & 0.0080 & 0.2962 & 1.3584 \\ \hline
7 & 0.0055 & 0.0100 & 0.5573 & 1.5576 \\ \hline
8 & 0.0049 & 0.0077 & 0.1043 & 0.8658 \\ \hline
9 & 0.6146 & 0.5902 & 0.6987 & 1.2436 \\ \hline
Average Error (m) & 0.0734 & 0.0737 & 0.3341 & 1.0246 \\ \hline
\end{tabular}
\end{table}

\begin{figure}[!t]
\centering
\includegraphics[width=0.7\columnwidth]{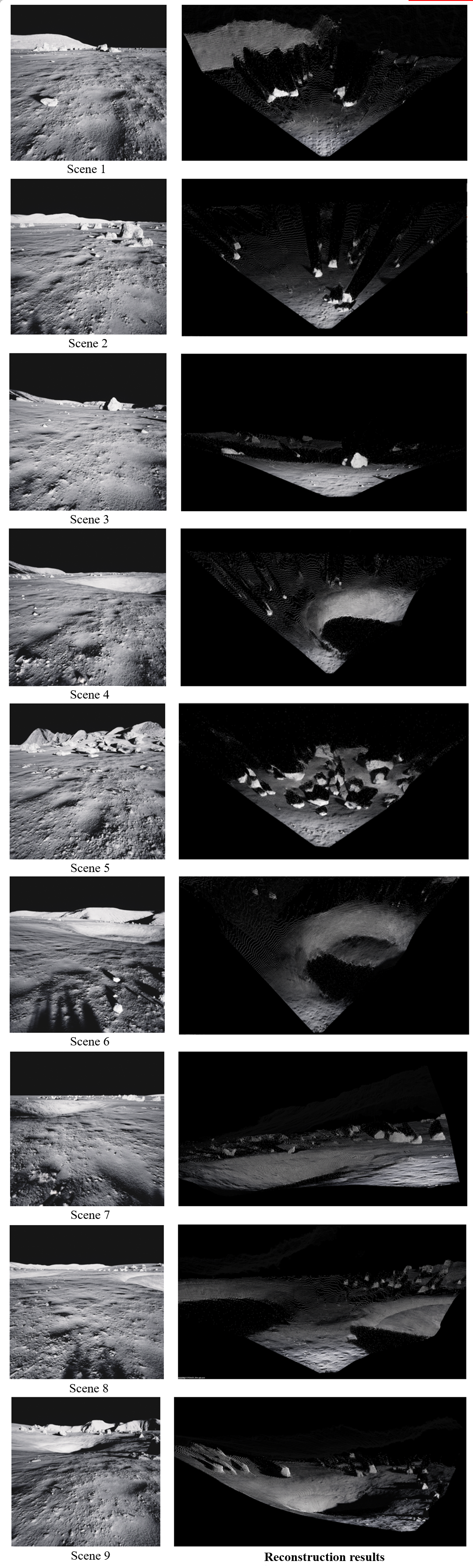}
\caption{Visualization of terrain reconstruction results from LuSNAR dataset.}
\label{figure_13}
\end{figure}

In the 3D reconstruction experiments, we chose the best model CREStereo in the stereo matching experiment to generate disparity images and obtain dense point cloud of the single station based on the camera's internal and external parameters. This paper uses chamfer distance as the quantitative criteria to compare the reconstructed point cloud with the point cloud restored by depth truth. The experimental data are consistent with the stereo matching experiment. Table~\ref{table_11} shows the accuracy evaluation results within the range of 5, 10, 20, and 50m.

Within the range of 5 meters, the average error of scenes 1-8 is 0.0057 meters, and 0.6146 meters in the complex scene 9; Within the range of 10 meters, the average error of 9 scenes is 0.0737 meters, and the average error of scenes 1-8 is 0.0092 meters. As shown in Fig.~\ref{figure_13}, the single station terrain reconstruction results based on CREStereo in scenarios 1-9 are shown. Scenes 1, 2, 3, and 5 show examples of scenes with different slopes and rock densities. Scenes 4, 6, 7, 8, and 9 show examples of different slopes and impact pits. Scene 8 has two impact pits. The results show that the reconstructed rock structures in the nine scenes are relatively complete, but due to the existence of perspective occlusion, some point clouds are missing; The reconstruction results of the impact crater are basically complete, and the edge geometry is clear. Taking the range of 50 meters as an example, the terrain reconstruction task is analyzed in combination with the topographic relief and surface feature richness. The results show that the relationship between reconstruction accuracy and terrain relief is the same as that of stereo matching task. Scenes 2, 5 and 8 with moderate terrain relief have higher accuracy, while scenes 3, 6 and 9 with high terrain relief have larger error. The accuracy of terrain reconstruction is negatively correlated with the richness of features, and the reconstruction accuracy decreases with the increasing richness of features. This may be due to the fact that within the range of 50 meters, scenes 1-3 are composed of lunar soil and a small amount of rocks, so it is less difficult to reconstruct the scene. With the increase of the types of features, impact pits and dense rocks will appear within tens of meters, making reconstruction more difficult.

\section{Conclusion}
This paper proposes a benchmark dataset LuSNAR for autonomous environmental perception and navigation on the lunar surface. This dataset collects diverse scene data based on a simulation engine and contains high-precision ground truth labels. We validate and evaluate this dataset using 2D/3D semantic segmentation, SLAM, and 3D reconstruction algorithms. The evaluation results show that the dataset provides high-quality label and ground truth information and can be used in the ground verification stage to validate and develop autonomous perception and navigation algorithms. We hope that LuSNAR will help promote the autonomy of lunar rover exploration and realize the universalization of intelligent technology in planetary scenes. In future work, we will expand the size of the LuSNAR dataset to cover more diverse lunar surface scenes. In addition, we will validate and establish baseline results on other tasks, such as 3D semantic reconstruction, absolute positioning relative to orbital data, and SLAM fusion of multi-source sensors. Through these works, we expect th LuSNAR dataset to provide more comprehensive data support for autonomous lunar rover exploration.

\section*{Acknowledgments}
The authors would like to thank the support from National Natural Science Foundation of China (No.42171445).

\bibliographystyle{IEEEtran}
\bibliography{reference}


 





\end{document}